\documentclass[10pt,twocolumn,letterpaper]{article} 

\usepackage{cvpr}
\usepackage{times}
\usepackage{epsfig}
\usepackage{graphicx}
\usepackage{amsmath}
\usepackage{amssymb}
\usepackage{multirow}
\usepackage{url}         
\usepackage{booktabs,tabularx}       
\usepackage{caption}
\usepackage{tablefootnote}

\newcommand{\myparagraph}[1]{\vspace{0.0em}\noindent\textbf{#1}}
\newcommand\Liqian[1]{\textcolor{blue}{}} %
\newcommand{\RN}[1]{%
  \textup{\uppercase\expandafter{\romannumeral#1}}%
}

\usepackage[pagebackref=true,breaklinks=true,letterpaper=true,colorlinks,bookmarks=false]{hyperref}

\cvprfinalcopy 

\def\cvprPaperID{1801} 

\begin{document}
\pagestyle{plain}
\title{Disentangled Person Image Generation}

\author{Liqian Ma$^{1}$ \quad Qianru Sun$^{2}\thanks{Corresponding author} $ \quad Stamatios Georgoulis$^{1}$ \\ 
Luc Van Gool$^{1,3}$ \quad Bernt Schiele$^{2}$  \quad Mario Fritz$^{2}$ \\
\\
$^{1}$KU-Leuven/PSI, Toyota Motor Europe (TRACE) \quad  $^{3}$ETH Zurich \\
  $^{2}$Max Planck Institute for Informatics, Saarland Informatics Campus \\
  {\texttt{\{liqian.ma, sgeorgou,
luc.vangool\}@esat.kuleuven.be}} \\
{\texttt{\{qsun, schiele, mfritz\}@mpi-inf.mpg.de}} 
}

\maketitle
\thispagestyle{empty}

\begin{abstract}
Generating novel, yet realistic, images of persons is a challenging task due to the complex interplay between the different image factors, such as the foreground, background and pose information. In this work, we aim at generating such images based on a novel, two-stage reconstruction pipeline that learns a disentangled representation of the aforementioned image factors and generates novel person images at the same time. First, a multi-branched reconstruction network is proposed to disentangle and encode the three factors into embedding features, which are then combined to re-compose the input image itself. Second, three corresponding mapping functions are learned in an adversarial manner in order to map Gaussian noise to the learned embedding feature space, for each factor, respectively. Using the proposed framework, we can manipulate the foreground, background and pose of the input image, and also sample new embedding features to generate such targeted manipulations, that provide more control over the generation process. Experiments on the Market-1501 and Deepfashion datasets show that our model does not only generate realistic person images with new foregrounds, backgrounds and poses, but also manipulates the generated factors and interpolates the in-between states. Another set of experiments on Market-1501 shows that our model can also be beneficial for the person re-identification task\footnote{Project page: \url{https://homes.esat.kuleuven.be/~liqianma/CVPR18_DPIG/index.html}}.
\end{abstract}


\vspace{-2mm}
\section{Introduction}
\label{sec:intro}

\begin{figure}[htp]
  \centering
  \includegraphics[width=1\linewidth]{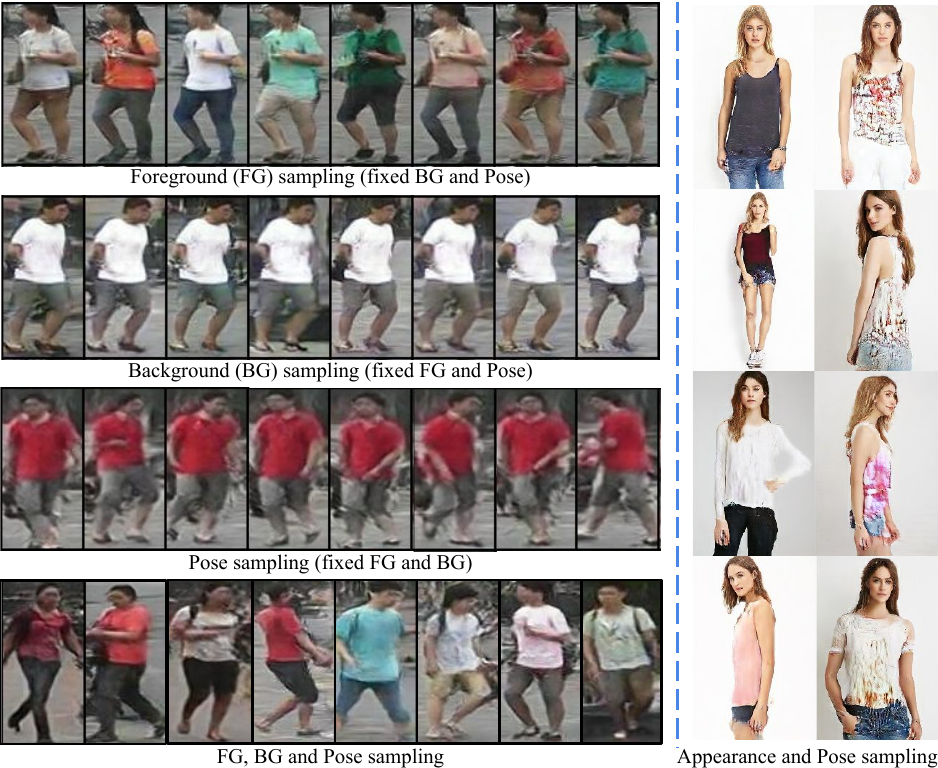}
  \vspace{-5mm}
  \caption{Left: image sampling results on Market-1501. Three factors, \ie foreground, background and pose, can be sampled independently (1st-3rd rows) and jointly (4th row). Right: similar joint sampling results on DeepFashion.
  This dataset contains almost no background, so we only disentangle the image into appearance and pose factors. 
  }
\label{fig:Paper_sampling_one_factor}
\vspace{-4mm}
\end{figure}

\vspace{-2mm}
The process of generating realistic-looking images of persons has several applications, like image editing, person re-identification (re-ID), inpainting or on-demand generated art for movie production.
The recent advent of image generation models, such as variational autoencoders (VAE) \cite{VAE}, generative adversarial networks (GANs) \cite{GAN} and autoregressive models (ARMs) (\eg PixelRNN \cite{van2016pixel}), has provided powerful tools towards this goal.
Several papers \cite{DCGAN, chen2016infogan, WGAN} have then exploited the ability of these networks to generate sharp images in order to synthesize realistic photos of faces and natural scenes.
Recently, Ma \etal \cite{PG2} proposed an architecture to synthesize novel person images in arbitrary poses given as input an image of that person and a new pose.

From an application perspective however, the user often wants to have more control over the generated images (\eg change the background, a person's appearance and clothing, or the viewpoint), which is something that existing methods are essentially uncapable of.
We go beyond these constraints and investigate how to generate novel person images with a specific user intention in mind (\ie foreground (FG), background (BG), pose manipulation).
The key idea is to explicitly guide the generation process by an appropriate representation of that intention.
Fig.~\ref{fig:Paper_sampling_one_factor} gives examples of the intended generated images.

To this end, we disentangle the input image into intermediate embedding features, \ie person images can be reduced to a composition of features of foreground, background, and pose. 
Compared to existing approaches, we rely on a different technique to generate new samples.
In particular, we aim at sampling from a standard distribution, \eg a Gaussian distribution, to first generate new embedding features and from them generate new images.
To achieve this, {\it fake} embedding features $\tilde{e}$ are learned in an adversarial manner to match the distribution of the {\it real} embedding features $e$, where the encoded features from the input image are treated as {\it real} whilst the ones generated from the Gaussian noise as {\it fake} (Fig.~\ref{fig:Paper_Framework_adver_decoder}).
Consequently, the newly sampled images come from learned {\it fake} embedding features $\tilde{e}$ rather than the original Gaussian noise as in the traditional GAN models. 
By doing so, the proposed technique enables us not only to sample a controllable input for the generator, but also to preserve the complexity of the composed images (\ie realistic person images).

To sum up, our full pipeline proceeds in two stages as shown in Fig.~\ref{fig:Paper_Framework_adver_decoder}.
At stage-\RN{1}, we use a person's image as input and disentangle the information into three main factors, namely foreground, background and pose. 
Each disentangled factor is modeled by embedding features through
a reconstruction network. 
At stage-\RN{2}, a mapping function is learned to map a Gaussian distribution to a feature embedding distribution.

Our contributions are: 
1) A new task of generating natural person images by disentangling the input into weakly correlated factors, namely foreground, background and pose. 
2) A two-stage framework to learn manipulatable embedding features for all three factors. 
In stage-\RN{1}, the encoder of the multi-branched reconstruction network serves conditional image generation tasks, whereas in stage-\RN{2} the mapping functions learned through adversarial training (\ie mapping noise $z$ to {\it fake} embedding features {\it emb}) serve sampling tasks (\ie the input is sampled from a standard Gaussian distribution).
3) A technique to match the distribution of {\it real} and {\it fake} embedding features through adversarial training, not bound to the image generation task.
4) An approach to generate new image pairs for person re-ID.
Sec.~\ref{sec:experiments} constructs a Virtual Market re-ID dataset by fixing foreground features and changing background features and pose keypoints to generate samples of one identity.  

\section{Related work}
\label{sec:related_work}

\begin{figure} 
\scriptsize
  \centering
  \includegraphics[width=1\linewidth]{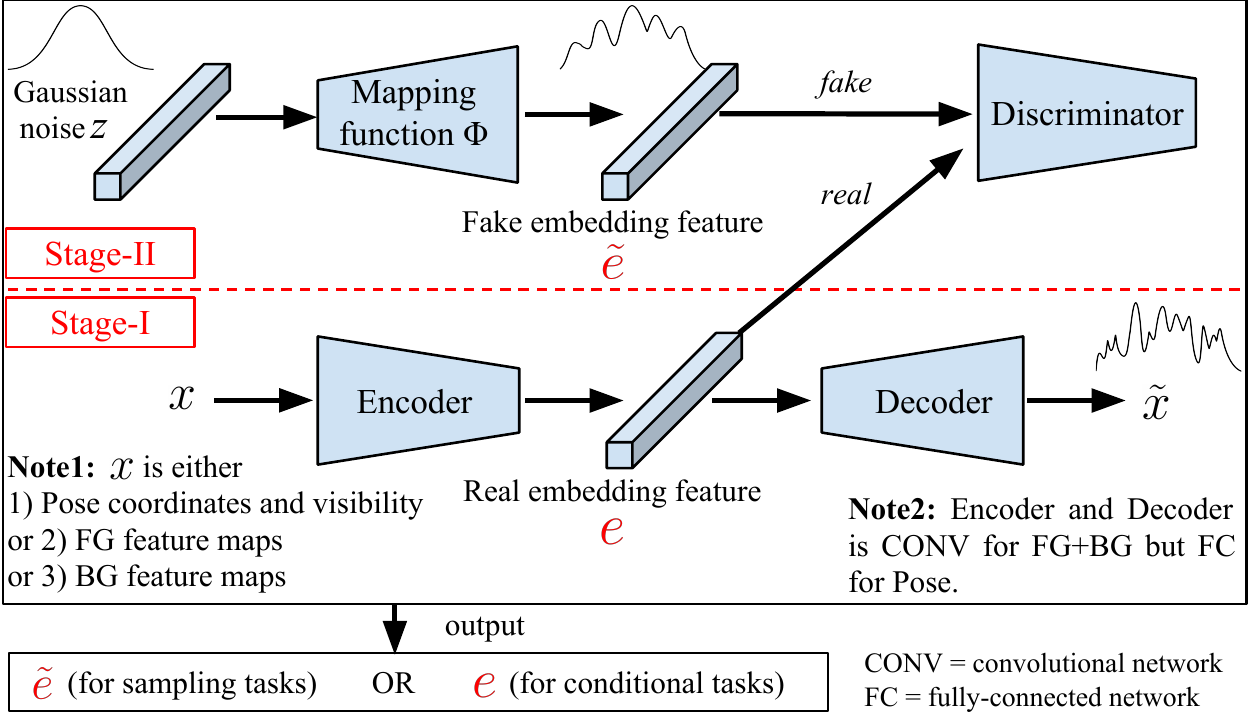}\\
\vspace{-2mm}
\caption{Our two-stage framework. In stage-\RN{1}, we use a reconstruction network to obtain the {\it real} embedding features $e$ for each factor, \ie foreground, background and pose. The architectural details of stage-\RN{1} are shown in Figure~\ref{fig:Paper_Framework_recons}. In stage-\RN{2}, we propose a novel, two-step mapping technique for adversarial embedding feature learning that first map Gaussian noise $z$ to intermediate embedding features $\tilde{e}$ then to the data $\tilde{x}$. We use the pre-trained encoder and decoder of stage-\RN{1} to guide the learning of mapping functions $\Phi$.}
\label{fig:Paper_Framework_adver_decoder}
\vspace{-3mm}
\end{figure}

\textbf{Image generation from noise.}
The ability of generative models, such as GANs \cite{GAN}, adversarial autoencoders (AAE) \cite{AAE}, VAEs \cite{VAE} and ARMs (\eg PixelRNN \cite{van2016pixel}), to synthesize realistic-looking, sharp images has led image generation research lately.
Traditional image generation works use GANs \cite{GAN} or VAEs \cite{VAE} to map a distribution generated by noise $z$ to the distribution of real data. 
Convolutional VAEs and AAEs \cite{AAE} have shown how to transform an auto-encoder into a generator, but in this case, it is rather difficult to train the mapping function for complex data distributions, such as person images (as also mentioned in ARAE-GAN \cite{ARAE-GAN}).
As such, traditional image generation methods are not optimal when it comes to the human body.
For example, Zheng \etal \cite{GAN_reid} directly adopted the DCGAN architecture \cite{DCGAN} to generate person images from noise, but as Fig.~\ref{fig:Paper_sampling}(b) shows, vanilla DCGAN leads to unrealistic results. 
Instead, we propose a two-step mapping technique in stage-\RN{2} to guide the learning, \ie $z \rightarrow e \rightarrow x$ (Fig.~\ref{fig:Paper_Framework_adver_decoder}).
Similar to \cite{ARAE-GAN}, we use a decoder to adversarially map the noise distribution to the feature embedding distribution learned by the reconstruction network.

\begin{figure*}
  \centering
  \includegraphics[width=1.00\linewidth]{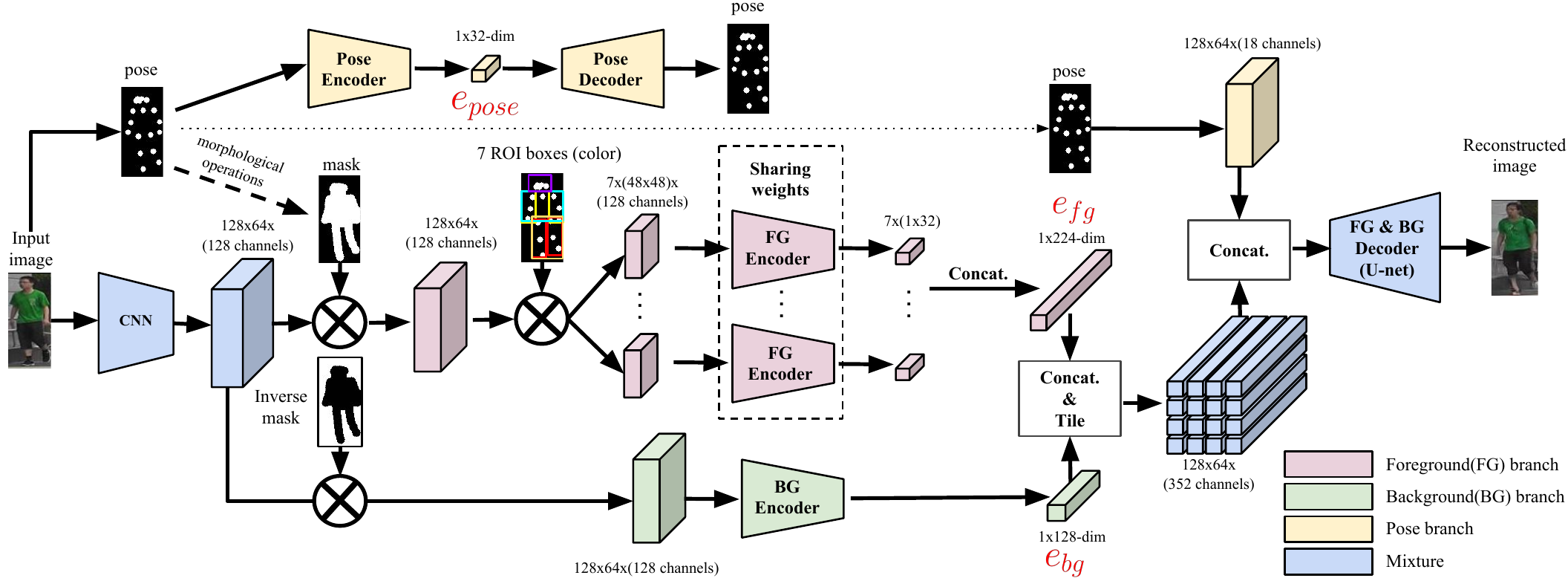}\\
\vspace{-2.5mm}
  \caption{Stage-\RN{1}: disentangled image reconstruction. This framework is composed of three branches: foreground, background and pose. Note that we use a fully-connected auto-encoder network to reconstruct the pose (incl. keypoint coordinates and visibility), so that we can decode the embedded pose features to obtain the heatmaps at the sampling phase.}
\vspace{-3mm}
\label{fig:Paper_Framework_recons}
\end{figure*}

\textbf{Conditional image generation.} 
Since the human body has a complex non-rigid structure with many degrees of freedom \cite{moeslund2006survey}, several works have used structure conditions to generate person images. 
Reed \etal in \cite{Reed-NIPS2016} proposed the Generative Adversarial What-Where Network that uses pose keypoints and text descriptions as condition, whereas in \cite{reed2016-techreport} they used an extension of PixelCNN in addition to conditioning on part keypoints, segmentation masks and text to generate images on the MPI Human Pose dataset, among others. 
Lassner \etal \cite{Lassner17ClothNet} generated full-body images of persons in clothing by conditioning on fine-grained body and clothing segments, \eg pose, shape or color. 
Zhao \etal \cite{Zhao-Arxiv16-Multiview} combined the strengths of GANs with variational inference to generate multi-view images of persons in clothing in a coarse-to-fine manner. 
Closer to our work, Ma \etal \cite{PG2} proposed to condition on image and pose keypoints to transfer the human pose in a flexible way. Facial landmarks can be transfered accordingly \cite{SUN2018}. Yet, their methods need the training set of aligned person image pairs which costs expensive human annotations. 
Most recently, Zhu \etal \cite{CycleGAN} proposed the CycleGAN that uses cycle consistency to achieve unpaired image-to-image translation between domains. 
They achieve compelling results in appearance changes but show little success in geometric changes.

Since images themselves contain abundant context information \cite{SunCVPR17}, some works have tried to tackle the problem in an unsupervised way.
Doersch \etal \cite{doersch2015unsupervised} explored the use of spatial context, \ie relative position between two neighboring patches in an image, as a supervisory signal for unsupervised visual representation learning. 
Noroozi \etal \cite{noroozi2016unsupervised} extended the task to a jigsaw puzzle solved by observing all the tiles simultaneously, which can reduce the ambiguity among these local patch pairs.
Lee \etal \cite{lee2017unsupervised} utilized context in an image generation task by inferring the spatial arrangement and generating the image at the same time.
We use the supervision in a different way. 
To extract pose-invariant appearance features, we arrange the body part feature embeddings according to the region-of-interest (ROI) bounding boxes obtained with pose keypoints. 
Then, we explicitly utilize these pose keypoints as structure information to select the necessary appearance features for each body part and generate the entire person image.

In general, this paper studies a different problem than these supervised or unsupervised approaches and tries to solve the disentangled person image generation task in an unpaired, self-supervised manner, by leveraging foreground, background and pose sampling at the same time, in order to gain more control over the generation process.

\textbf{Disentangled image generation.}
Few papers have already tried to work towards this direction by learning a disentangled representation of the input image. 
Chen \etal \cite{chen2016infogan} proposed InfoGAN, an extension to GANs, to learn disentangled representations using mutual information in an unsupervised manner, like writing styles from digit shapes on the MNIST dataset, pose from lighting of 3D rendered images, and background digits from the central digit on the SVHN dataset.
Cheung \etal \cite{cheung2014discovering} added a cross-covariance penalty in a semi-supervised autoencoder architecture in order to disentangle factors, like hand-writing style for digits and subject identity in faces. 
Tran \etal \cite{tran2017disentangled} proposed DR-GAN to learn both a generative and a discriminative representation from one or multiple face images to synthesize identity-preserving faces at target poses.
In contrast, our method gives an explicit representation of the main 3 axis of variation (foreground, background, pose). 
Moreover, training is facilitated without a need for expensive identity annotations - which is not readily available at scale.

\vspace{-1mm}
\section{Method}
\label{sec:method}
\vspace{-2mm}

Our goal is to disentangle the appearance and structure factors in person images, so that we can manipulate the foreground, background and pose separately. 
To achieve this, we propose a two-stage pipeline shown in Fig.~\ref{fig:Paper_Framework_adver_decoder}. 
In stage-\RN{1}, we disentangle the foreground, background and pose factors using a reconstruction network in a divide-and-conquer manner. 
In particular, we reconstruct person images by first disentangling into intermediate embedding features of the three factors, then recover the input image by decoding these features. 
In stage-\RN{2}, we treat these features as {\it real} to learn mapping functions $\Phi$ for mapping a Gaussian distribution 
to the embedding feature distribution adversarially.

\subsection{Stage-\RN{1}: Disentangled image reconstruction}
\label{sec:method_stageI}
At stage-\RN{1}, we propose a multi-branched reconstruction architecture to disentangle the foreground, background and pose factors as shown in Fig.~\ref{fig:Paper_Framework_recons}. 
Note that, to obtain the pose heatmaps and the coarse pose mask we adopt the same procedure as in \cite{PG2}, but we instead use them to guide the information flow in our multi-branched network.

\myparagraph{Foreground branch.} 
To separate the foreground and background information, we apply the coarse pose mask to the feature maps instead of the input image directly.
By doing so, we can alleviate the inaccuracies of the coarse pose mask. 
Then, in order to further disentangle the foreground from the pose information, we encode pose invariant features with 7 Body Regions-Of-Interest instead of the whole image similar to \cite{RegionReid2017}. 
Specifically, for each ROI we extract the feature maps resized to 48$\times$48 and pass them into the weight sharing foreground encoder to increase the learning efficiency. 
Finally, the encoded 7 body ROI embedding features are concatenated into a 224D feature vector. 
Later, we use BodyROI7 to denote our model which uses 7 body ROIs to extract foreground embedding features, and use WholeBody to denote our model that extracts foreground embedding features from the whole feature maps directly instead of extracting and resizing the ROI feature map.

\myparagraph{Background branch.} 
For the background branch, we apply the inverse pose mask to get the background feature maps and pass them into the background encoder to obtain a 128-dim embedding feature. 
Then, the foreground and background features are concatenated and tiled into 128$\times$64$\times$352 appearance feature maps.

\myparagraph{Pose branch.} 
For the pose branch, we concatenate the 18-channel heatmaps with the appearance feature maps and pass them into the a ``U-Net''-based architecture~\cite{U-net}, \ie, convolutional autoencoder with skip connections, to generate the final person image following PG$^2$ (G1+D) \cite{PG2}. 
Here, the combination of appearance and pose imposes a strong explicit disentangling constraint that forces the network to learn how to use pose structure information to select the useful appearance information for each pixel.
For pose sampling, we use an extra fully-connected network to reconstruct the pose information, so that we can decode the embedded pose features to obtain the heatmaps.
Since some body regions may be unseen due to occlusions, we introduce a visibility variable $\alpha_{i} \in \{0,1\}, i=1,...,18$ to represent the visibility state of each pose keypoint. 
Now, the pose information can be represented by a 54-dim vector (36-dim keypoint coordinates $\gamma$ and 18-dim keypoint visibility $\alpha$).

\begin{figure*}[htp]
  \centering
  \includegraphics[width=0.93\linewidth]{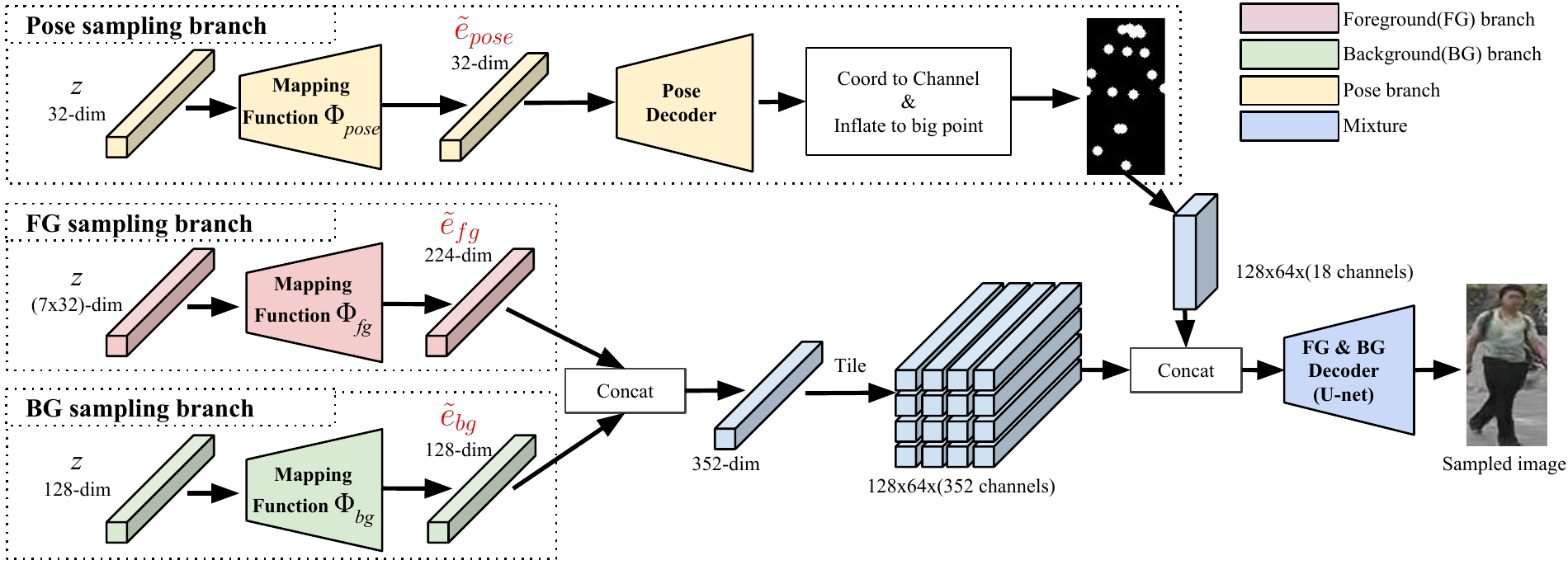}\\
\vspace{-1mm}
  \caption{Sampling phase: Sample foreground, background and pose from Gaussian noise to compose new person images. }
\vspace{-2mm}
\label{fig:Paper_Framework_sampling}
\end{figure*}

\subsection{Stage-\RN{2}: Embedding feature mapping}
Images can be represented by a low-dimensional, continuous feature embedding space. 
In particular, in \cite{semidefinite_Manifold, LLE_Manifold, LCC, dollar2007learning} it has been shown that they lie on or near a low-dimensional manifold of the original high-dimensional space. 
Therefore, the distribution of this feature embedding space should be more continuous and easier to learn compared to the real data distribution. 
Some works \cite{Zhang-stackGAN, gupta2016cross, romero2014fitnets} have then attempted to use the intermediate feature representations of a pre-trained DNN to guide another DNN.
Inspired by these ideas, we propose a two-step mapping technique as illustrated in Fig.~\ref{fig:Paper_Framework_adver_decoder}. 
Instead of directly learning to decode Gaussian noise to the image space, we first learn a mapping function $\Phi$ that maps a Gaussian space $\textbf{Z}$ into a continuous feature embedding space $\textbf{E}$, and then use the pre-trained decoder to map the feature embedding space $\textbf{E}$ into the real image space $\textbf{X}$. 
The encoder learned in stage-\RN{1} encodes the FG, BG and Pose factors $x$ into low-dimensional {\it real} embedding features $e$. 
Then, we treat the features mapped from Gaussian noise $z$ as {\it fake} embedding features $\tilde{e}$ and learn the mapping function $\Phi$ adversarially. 
In this way, we can sample {\it fake} embedding features from noise and then map them back to images using the decoder learned in stage-\RN{1}.
The proposed two-step mapping technique is easy to train in a piecewise style and most importantly can be useful for other image generation applications.

\subsection{Person image sampling}
As explained, each image factor can not only be encoded from the input information, but also be sampled from Gaussian noise.
As to the latter, to sample a new foreground, background or pose, we combine the decoders learned in stage-\RN{1} and mapping functions learned in stage-\RN{2} to construct a $z \rightarrow \tilde{e} \rightarrow \tilde{x}$ sampling pipeline (Fig.~\ref{fig:Paper_Framework_sampling}). 
Note that, for foreground and background sampling the decoder is a convolutional ``U-net''-based architecture, while for pose sampling the decoder is a fully-connected one.
Our experiments show that our framework performs well when used in both a conditional and an unconditional way.

\subsection{Network architecture}
Here, we describe the proposed architecture. 
For both stages, we use residual blocks to make the training easier. 
All convolution layers consist of 3$\times$3 filters and the number of filters increases linearly with each block. 
All fully-connected layers consist of 512-dim, except for the bottleneck layers. 
We apply rectified linear units (ReLU) to each layer, except for the bottleneck and the output layers.

For the foreground and background branches in stage-\RN{1}, the input image is fed into a convolutional residual block and the pose mask is used to extract the foreground and background feature maps. 
Then, the masked foreground and background feature maps are passed into an encoder consisting of $N$ convolutional residual blocks, respectively, where $N$ depends on the size of the input. 
Similar to \cite{PG2}, each residual block consists of two convolution layers with stride=1, followed by one sub-sampling convolution layer with stride=2, except for the last block. 
For the decoder, an ``U-Net''-based architecture \cite{U-net} is used with $N$ convolutional residual blocks before and after the bottlenecks, respectively, following PG$^2$ (G1+D) \cite{PG2}.

For pose reconstruction, we use an auto-encoder architecture where both encoder and decoder consist of 4 fully-connected residual blocks with 32-dim bottleneck layers. 
As in \cite{huang2016densely}, we use a densely-connected-like architecture, \ie each residual block consists of two fully-connected layers. 

For each mapping function in stage-\RN{2}, we use a fully-connected network consisting of 4 fully-connected residual blocks to map $K$-dim Gaussian noise $z$ to $K$-dim embedding features $e$. 
For the discriminator, we adopt a fully-connected network with 4 fully-connected layers.

\subsection{Optimization strategy}
The training procedures of stage-\RN{1} and stage-\RN{2} are separated, since the mapping functions $\Phi_{\textit{fg}}$, $\Phi_{\textit{bg}}$ and $\Phi_{\textit{pose}}$ in stage-\RN{2} can be trained in a piecewise style. 
In stage-\RN{1}, we use both L1 and adversarial loss to optimize the image (\ie foreground and background) reconstruction network. This choice is known to result in sharper and more realistic images. 
In particular, we use $G_1$ and $D_1$ to denote the image reconstruction network  and the corresponding discriminator in stage-\RN{1}.
The overall losses for $G_1$ and $D_1$ are as follows,
\begin{align}
    {\cal{L}}^{D_1}_{R} =& {\mathbb{E}}_{x \sim p_{\textit{data}}(x)}\big[\log{D_1(x)}\big] + \nonumber \\  
    &{\mathbb{E}}_{x \sim p_{\textit{data}}(x)}\big[\log{(1-D_1(G_1(x,h)))}\big], 
    \label{eq:recon_D_loss} \\
    {\cal{L}}^{G_1}_{R} =& {\mathbb{E}}_{x \sim p_{\textit{data}}(x)}\big[\log{(D_1(G_1(x,h)))}\big] + \nonumber \\ 
    &\lambda \| (G_1(x,h) - x) \|_1, 
    \label{eq:recon_G_loss}
\end{align} 
where $x$ denotes the person image, $h$ denotes the pose heatmaps, and $\lambda$ is the weight of L1 loss controlling how close the reconstruction looks like to the input image at low frequencies. 
For pose reconstruction, we use the L2 loss to reconstruct the input pose information including keypoint coordinates $\gamma$ and visibility $\alpha$ mentioned in Sec.~\ref{sec:method_stageI},
\begin{equation}
    {\cal{L}}^{\textit{Pose}}_{R} = {\mathbb{E}}_{(\gamma,\alpha) \sim p_{\textit{data}}(\gamma,\alpha)} \| (G_1(\gamma,\alpha) - (\gamma,\alpha) \|^2_2, 
    \label{eq:recon_pose_loss}
\end{equation}

After training the reconstruction network in stage-\RN{1}, we fix it and use the Wasserstein GAN \cite{WGAN} loss to optimize the fully-connected network of mapping functions in stage-\RN{2}. We use $\Phi$ and $D_2$ to denote the mapping functions (incl. $\Phi_{\textit{fg}}$, $\Phi_{\textit{bg}}$ and $\Phi_{\textit{pose}}$) and the corresponding discriminators in stage-\RN{2}.
The overall losses for $\Phi$ and $D_2$ are as follows,
\begin{align}
    {\cal{L}}^{D_2}_{M} = & {\mathbb{E}}_{e \sim p_{\textit{emb}}(e)}\big[{D_2}(e)\big] - {\mathbb{E}}_{z \sim p_{z}(z)}\big[{D_2}(\Phi(z))\big], 
    \label{eq:emb_G_loss} \\
    {\cal{L}}^{\Phi}_{M} = & {\mathbb{E}}_{z \sim p_{z}(z)}\big[{D_2}(\Phi(z))\big], 
    \label{eq:emb_G_loss}
\end{align} 
where $e$ denotes the embedding features extracted from the reconstruction network in stage-\RN{1}, $z$ denotes the Gaussian noise.
Note that, we also tried the vanilla GAN loss but suffered a model collapse. 
For adversarial training, we optimize the discriminator and generator alternatively.

\section{Experiments}
\vspace{-0.5mm}
\label{sec:experiments}
The proposed pipeline enables many applications, incl. image manipulation, pose-guided person image generation, image interpolation, image sampling and person re-ID.\footnote{More generated results, parameters of our network architecture and training details are given in the supplementary material.}

\subsection{Datasets and metrics}
\vspace{-1mm}

Our main experiments use the challenging re-ID dataset Market-1501 \cite{Market1501}, containing 32,668 images of 1,501 persons captured from six disjoint surveillance cameras. 
All images are resized to 128$\times$64 pixels. 
We use the same train/test split (12,936/19,732) as in \cite{Market1501}, but use all the images in the train set for training without any identity label. 
For pose-guided person image generation, we randomly select 12,800 pairs in the test set for testing, following \cite{PG2}. 
For re-ID, we follow the same testing protocol as in \cite{Market1501}.

We also experiment with a high-resolution dataset, namely DeepFashion (In-shop Clothes Retrieval Benchmark) \cite{DeepFashion}, that consists of 52,712 in-shop clothes images and ~200,000 cross-pose/scale pairs. 
Following \cite{PG2}, we use the up-body person images and filter out failure cases in pose estimation for both training and testing. 
Thus, we have 15,079 training images and 7,996 testing images. 
We also randomly select 12,800 pairs from the test set for pose-guided person image generation testing.

\myparagraph{Implementation details.}
For Market-1501, our method is applied to disentangle the image into three factors: foreground, background and pose.
We set the number of convolutional residual blocks $N=5$ for foreground and background encoders and decoders. 
For DeepFashion, since it contains almost no background, we disentangle the images into only two factors: appearance and pose.
We set the number of convolution blocks $N=7$ for the foreground encoder and decoder.
On both datasets, we do a left-right flip data augmentation.
For the pose keypoints and mask extraction, we use the same procedure as \cite{PG2}. 

\subsection{Image manipulation}
\vspace{-0.5mm}
As explained, a person's image can be disentangled into three factors: FG, BG and Pose. 
Each factor can then be generated either from a Gaussian signal (sampling) or conditioned on input data, namely image and pose (conditioning). 
The conditional case contains at least one other factor sampled from Gaussian signals.
In Fig.~\ref{fig:Paper_sampling_one_factor}, the left-top3 rows show examples with one-factor sampling and two-factor conditioning for FG, BG and Pose on Market-1501, respectively. 
Our framework successfully manipulates each intended factor while keeping the others unchanged. 
In the first row, we sample foreground with $z_{\textit{fg}} \rightarrow \tilde{e}_{\textit{fg}}$ and condition background and pose with $x \rightarrow e$, so that different cloth colors, styles and hair styles can be generated while the pose and background stay mostly the same. 
Similarly, we can manipulate the background and pose independently as shown in the left-second/third row.
The left-last row shows a sampling example without any conditioning. 
In this way, we can sample novel person images from noise and still generate realistic images compared to vanilla VAE and DCGAN as shown in Sec.~\ref{sec:sampling_comparison}. 
Finally, on the right rows we show that our method can also sample 256$\times$256 images with realistic cloth and hair details on DeepFashion.

\begin{figure*} [htbp]
\scriptsize
\centering
  \centering
  \includegraphics[width=0.99\linewidth]{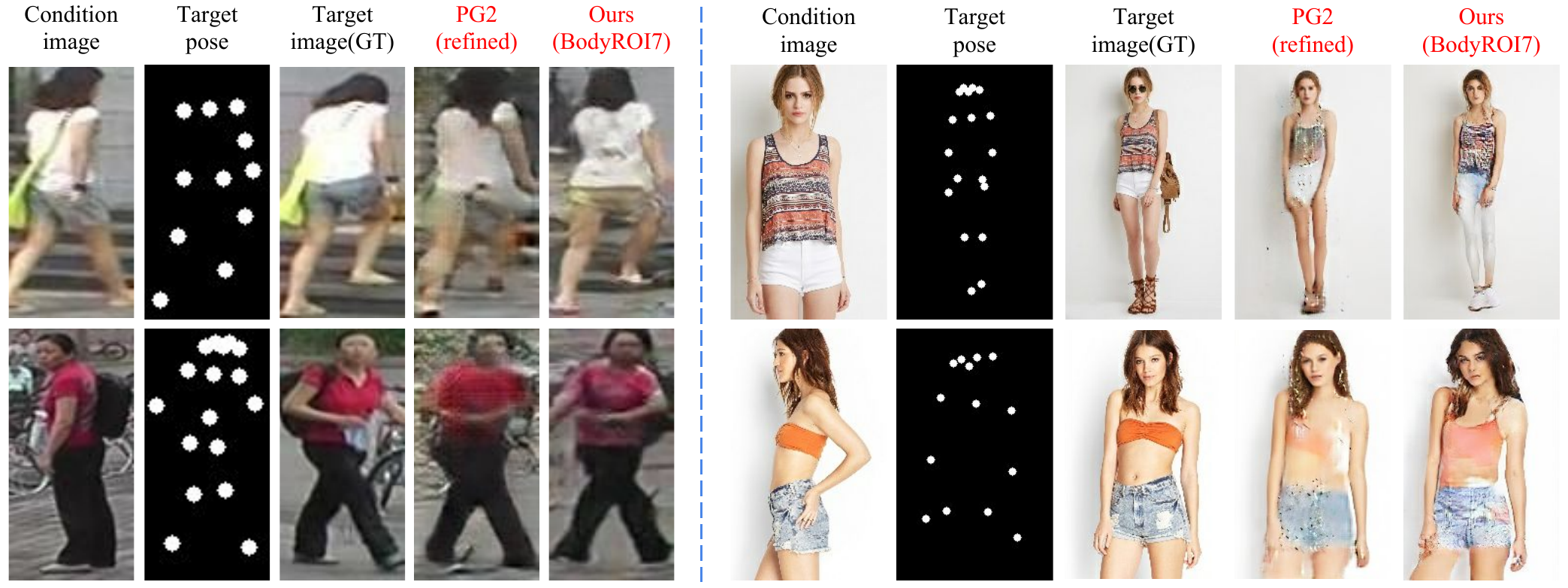}\\
\caption{Comparison to PG$^2$. Left: results on Market-1501. Right: results on DeepFashion. Zoom in for details.}
\label{fig:Paper_comparison_PG2}
\end{figure*}

\begin{figure*} [htp]
\scriptsize
\centering
\begin{minipage}{0.32\textwidth}
\hspace{-0.2cm}
  \centering
  \includegraphics[width=0.9\linewidth]{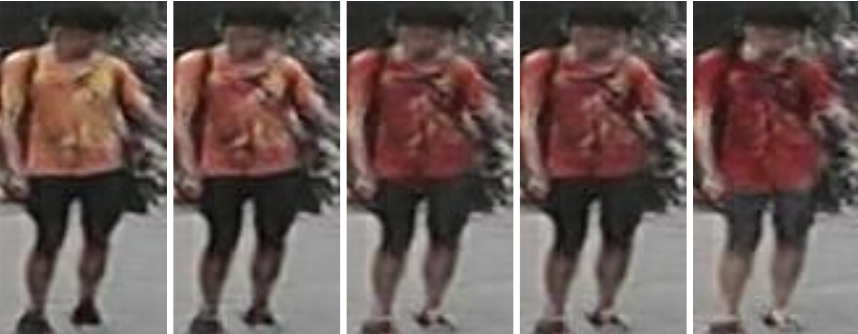}\\
(a) Foreground interpolation
\end{minipage}
\hfill
\begin{minipage}{0.32\textwidth}
  \centering
  \includegraphics[width=0.9\linewidth]{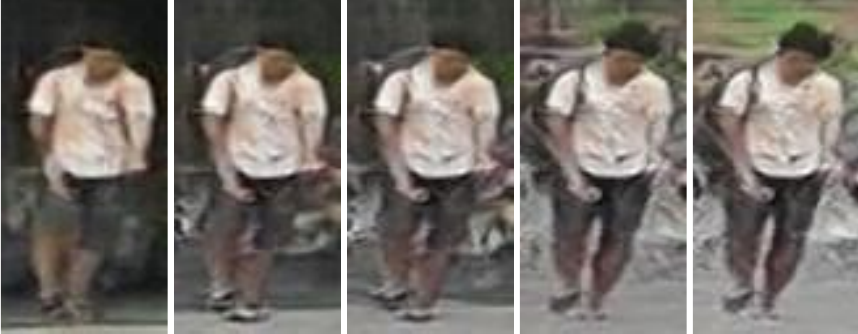}\\
(b) Background interpolation
\end{minipage}
\hfill
\begin{minipage}{0.32\textwidth}
  \centering
  \includegraphics[width=0.9\linewidth]{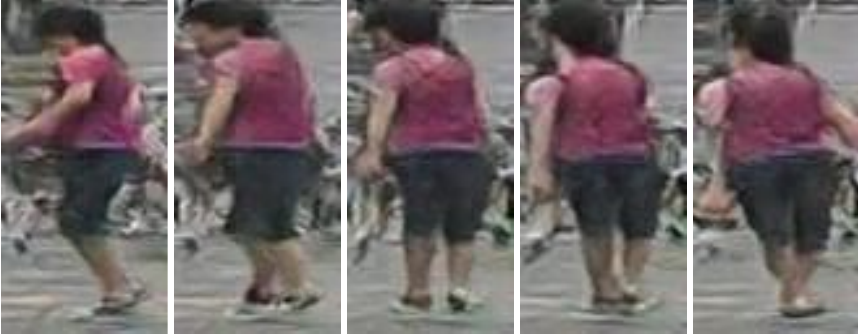}\\
(c) Pose interpolation
\end{minipage}
\hfill
\begin{minipage}{0.99\textwidth}
  \centering
  \includegraphics[width=0.9\linewidth]{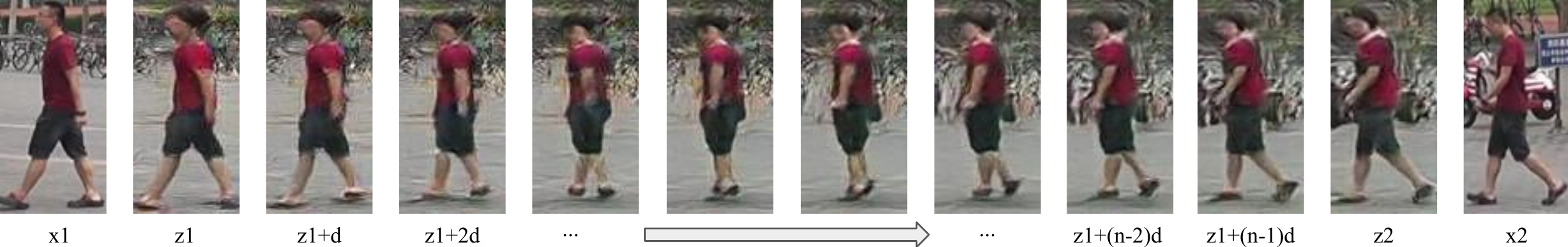}\\
(d) Inverse interpolation between two images.
\end{minipage}
  \caption{Factor interpolation. (a)(b)(c) We randomly select two Gaussian codes $z_1$ and $z_2$ and interpolate codes between $z_1$ and $z_2$ linearly; we then generate the interpolated images accordingly. (d) We invert an image pair first to embedding features $e_1$ and $e_2$, then to Gaussian codes $z_1$ and $z_2$. We then follow the same procedure as in (a)(b)(c).}
\label{fig:Paper_interpolate}
\end{figure*}

\subsection{Pose-guided person image generation}
\vspace{-0.5mm}
We compare our method with PG$^2$ \cite{PG2} on pose-conditional person image generation. 
Unlike PG$^2$, our method does not need paired training images.
As shown in Fig.~\ref{fig:Paper_comparison_PG2}, our method can generate more realistic details and less artifacts. 
Especially, the arms and legs are better shaped on both datasets, and the hair details are more clear on DeepFashion.
This is in agreement with the Inception Score (IS) and mask Inception Score (mask-IS) in Table~\ref{tab:compare_to_PG2}. 
The SSIM score of our method is lower than PG$^2$ mainly for two reasons.
1) In stage-\RN{1}, there are no skip-connections between encoder and decoder, and as such our method has to generate images from compressed embedding features instead of pixel level transforms like in PG$^2$, which is a harder task.
2) Our method generates sharper images which might decrease the SSIM score, as also observed in \cite{PG2,Johnson-ECCV16-Superres,Shi-CVPR16-Superres}. 

\begin{table}
\centering
\footnotesize
\begin{tabular*}{8.5cm}
{@{\extracolsep{\fill}} l c c c c c c }
\toprule 
& \multicolumn{2}{c}{DeepFashion} & \multicolumn{4}{c}{Market-1501} \\
\cmidrule{2-3} \cmidrule{4-7}
Model & SSIM & IS & SSIM & IS & mask-SSIM & mask-IS \\
\midrule[0.6pt]	
	PG$^2$\cite{PG2} & 0.762 & 3.090 &0.253 & 3.460 & 0.792 & 3.435 \\
    Ours & 0.614 & 3.228 & 0.099 & 3.483 & 0.614 & 3.491 \\
\bottomrule[1pt]
\end{tabular*}
\vspace{-0.2cm}
\caption{Quantitative evaluation. Higher scores are better.} 
\vspace{-0.3cm}
\label{tab:compare_to_PG2}
\end{table}

\begin{figure*} [htp]
\scriptsize
\begin{minipage}{0.33\textwidth}
\centering
  \centering
  \includegraphics[width=0.95\linewidth]{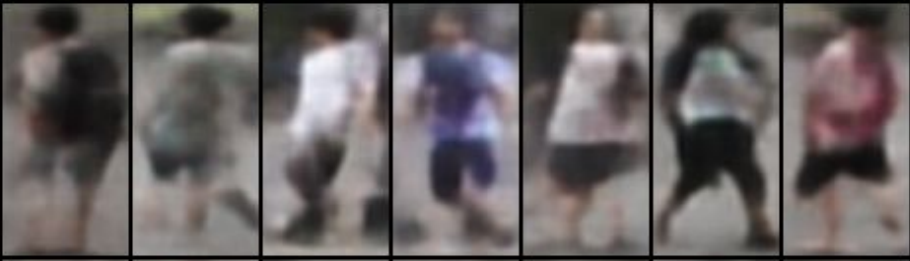}\\
(a) VAE~\cite{VAE}
\vspace{1mm}
\end{minipage}
\hfill
\begin{minipage}{0.33\textwidth}
\centering
  \centering
  \includegraphics[width=0.95\linewidth]{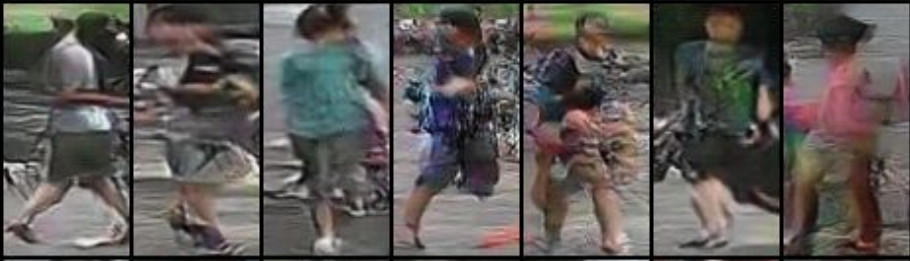}\\
(b) DCGAN~\cite{DCGAN}
\vspace{1mm}
\end{minipage}
\hfill
\begin{minipage}{0.33\textwidth}
\centering
  \centering
  \includegraphics[width=0.95\linewidth]{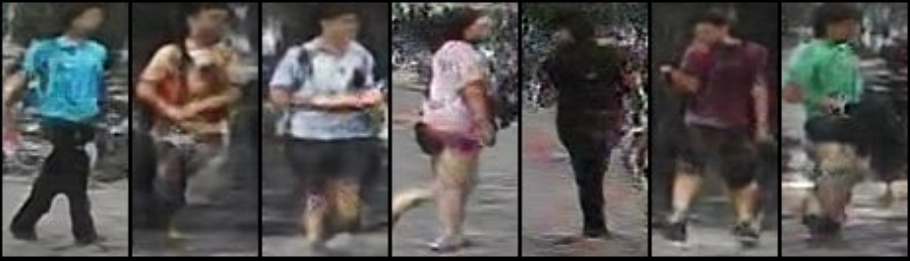}\\
(c) Ours - Whole Body
\vspace{1mm}
\end{minipage}
\hfill
\begin{minipage}{0.33\textwidth}
\centering
  \centering
  \includegraphics[width=0.95\linewidth]{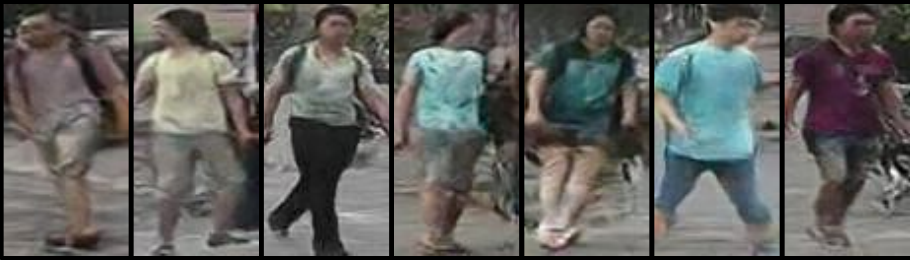}\\
(d) Ours - BodyROI7
\vspace{-0.5mm}
\end{minipage}
\hfill
\begin{minipage}{0.33\textwidth}
\centering
  \centering
  \includegraphics[width=0.95\linewidth]{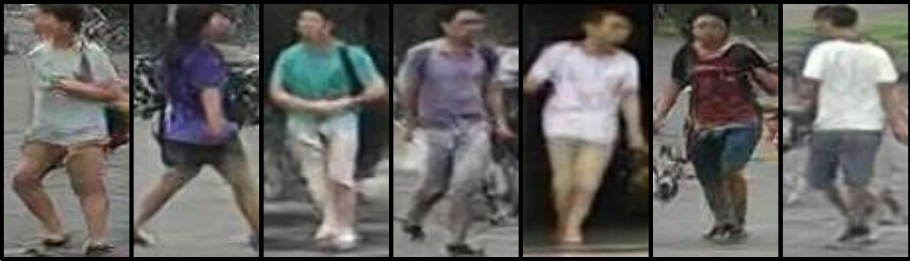}\\
(e) Ours - BodyROI7 with real pose from training set
\vspace{-0.5mm}
\end{minipage}
\hfill
\begin{minipage}{0.33\textwidth}
\centering
  \centering
  \includegraphics[width=0.95\linewidth]{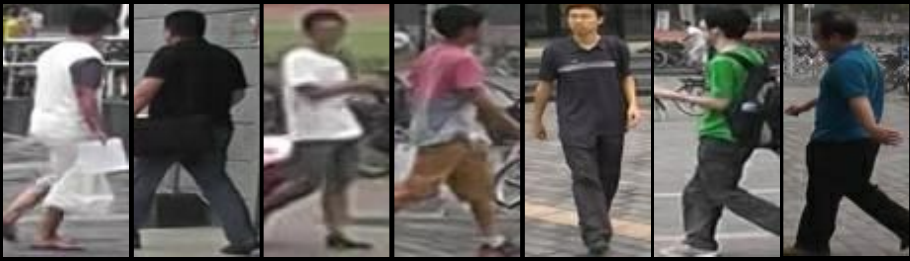}\\
(f) Real data
\vspace{-0.5mm}
\end{minipage}
\vspace{-0.5mm}
\caption{Sampling results comparison. From left to right and from top to bottom: (a) VAE~\cite{VAE} (b) DCGAN~\cite{DCGAN} (c) Ours - Whole Body (d) Ours - BodyROI7 (e) Ours - BodyROI7 with real pose from training set (f) Real data.}
\label{fig:Paper_sampling}
\end{figure*}

\subsection{Image interpolation}
\vspace{-0.5mm}
\label{sec:exp_interpolate}
Interpolation is possible for sampled and real images. 

\myparagraph{Sampling interpolation.}
For sampling interpolation, we directly interpolate in Gaussian space and generate images in a $z \rightarrow \tilde{e} \rightarrow \tilde{x}$ manner. 
We first interpolate linearly between two Gaussian codes $z_1$ and $z_2$ to obtain intermediate codes $z_i$, which in turn are mapped into embedding features $\tilde{e}_i$ using the learned mapping functions. 
The person's image is then generated from the embedding features $\tilde{e}_i$.
As Fig.~\ref{fig:Paper_interpolate}(a)(b)(c) shows, our method can smoothly interpolate each factor in Gaussian space separately, hence: 
1) our method can learn foreground, background and pose encoders in a disentangled way; 
2) these can map real high-dimensional data distributions into continuous low-dimensional feature embedding distributions; 
3) the mappings trained adversarially can map Gaussian to feature embedding distributions; 
4) the decoder can map feature embedding distributions back to real data distributions.

\myparagraph{Inverse interpolation}
To interpolate between real data (incl. image and pose keypoints), we proceed in 3 steps. 
1) $x \rightarrow e$: Use the learned encoders to encode real data $x$ into embedding features $e$. 
2) $e \rightarrow z$: Use gradient-based minimization \cite{inverse} to find the corresponding Gaussian codes $z$. 
3) $z \rightarrow \tilde{e} \rightarrow \tilde{x}$: Interpolate linearly between two Gaussian codes, then map intermediate codes into embedding features - using the learned mapping functions - to generate the person images.
As shown in Fig.~\ref{fig:Paper_interpolate}, our method interpolates reasonable frames between the input pair showing a person with different poses. 
The result shows realistic intermediate states and can be used to predict potential behaviors.

\begin{figure}[htp]
  \centering
  \includegraphics[width=0.99\linewidth]{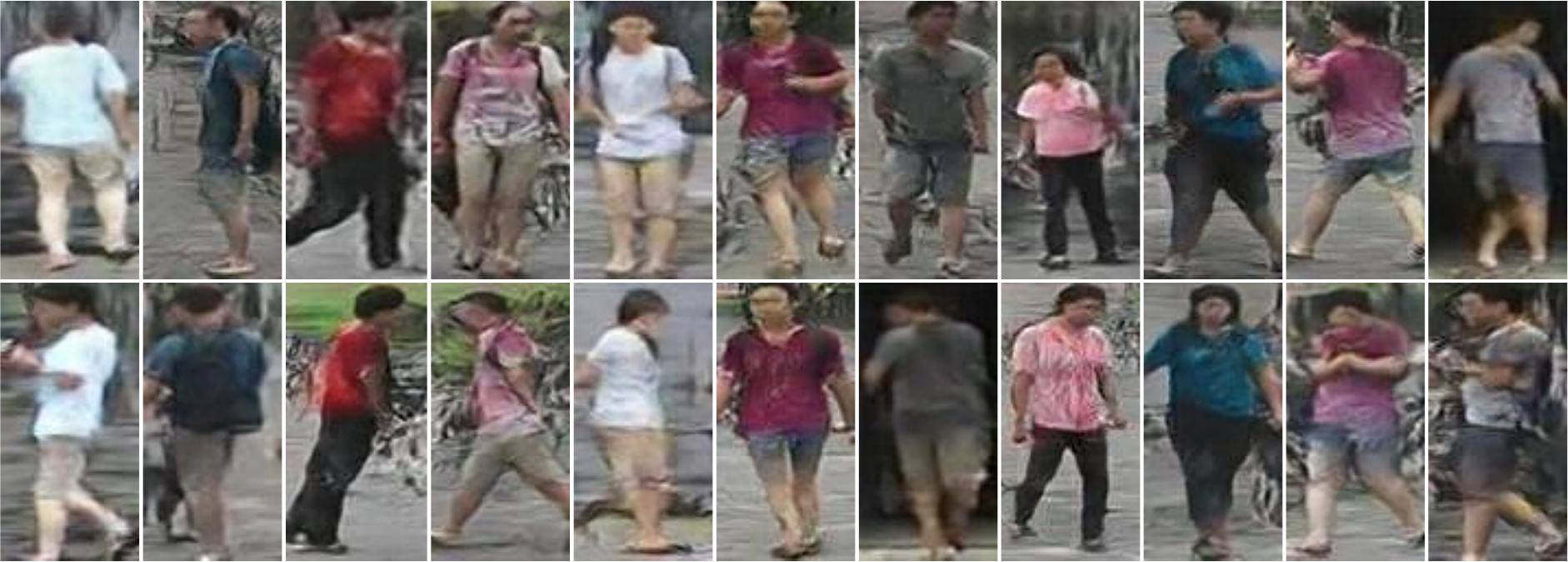}\\
\vspace{-1mm}
  \caption{Virtual identities for re-ID model training. Each column contains a pair of images of one identity (one FG). BG and Pose are randomly selected from training data.}
\label{fig:Paper_virtual_market}
\end{figure}

\subsection{Sampling results comparison}
\vspace{-0.5mm}
\label{sec:sampling_comparison}

In this experiment, we compare sampling results from our method and baseline models, \ie VAE~\cite{VAE} and DCGAN~\cite{DCGAN}. 
As illustrated in Fig.~\ref{fig:Paper_sampling}, VAE generates blurry images and DCGAN sharp but unrealistic person images. 
In contrast, our model generates more realistic images (see Fig.~\ref{fig:Paper_sampling}(c)(d)(e)). 
By comparing (d) and (c), we observe that our model using body ROI generates more sharp and realistic images whose colors on each body part are more natural. 
A similar tendency can be observed for re-ID.
By comparing (e) and (d), we see that when sampling foreground and background but using the real pose keypoints randomly selected from the training data, we generate better results. 
Therefore, we use this setting in (e) to sample virtual data for the following re-ID experiment.

\begin{table}
\centering
\footnotesize
\begin{tabular*}{8.4cm}
{@{\extracolsep{\fill}} l c c c c }
\toprule 
Model & Training data & Rank-1 & mAP \\
\midrule[0.6pt]	
	Bow \cite{Market1501} & Market & 0.344 & 0.141 \\
	Bow* \cite{Market1501} & Market & 0.358 & 0.148 \\
    LOMO* \cite{LOMO} & / & 0.272 & 0.08 \\
    WholeBody feature (Ours) & Market & 0.307 & 0.100 \\
    BodyROI7 feature (Ours) & Market & 0.338 & 0.107 \\
    BodyROI7 feature PCA (Ours) & Market & 0.355 & 0.114 \\
\midrule[1pt]
    Res50* \cite{PUL} & CUHK03 (labeled) & 0.300 & 0.115 \\
    Res50* \cite{PUL} & Duke (labeled) & 0.361 & 0.142 \\
    Res50 & VM & 0.338 & 0.134 \\
    Res50+PUL & VM+Market & 0.369 & 0.156 \\
    Res50+PUL+KISSME & VM+Market & 0.375 & 0.154 \\
\bottomrule[1pt]
\end{tabular*}
\vspace{-1.5mm}
\caption{Re-ID results on Market-1501. Top: using embedding features. Bottom: using VM and Market-1501 dataset without labels. Higher scores are better.
*Results are reported in \cite{PUL}. / means that hand-crafted feature extractor LOMO does not require training data.
}
\vspace{-2.5mm}
\label{tab:re-ID}
\end{table}

\subsection{Person re-identification}
\vspace{-1mm}
Person re-ID associates images of the same person across views or time.
Given the query person image, re-ID is expected to provide matching images of the same identity.
We propose to use the re-ID performance as a quantitative metric for our generation approach. 
We adopt the re-ID model in \cite{PUL} and use rank-1 matching rate and mean Average Precision (mAP) following \cite{Market1501}. 
We show that our approach can be evaluated in two ways: (1) use FG features extracted in stage-\RN{1} for re-ID; (2) generate virtual image pairs to train re-ID model. The virtual market data is denoted as ``VM'' generated with our BodyROI7 model. 
Note that, CUHK03 \cite{CUHK03} and Duke \cite{GAN_reid} datasets are used with identity labels, while Market-1501 and VM datasets are used with no labels.

\myparagraph{Using embedding features.}
We use the FG encoder to extract the features for re-ID and use the re-ID performance to evaluate the reconstruction network in stage-\RN{1}.
Intuitively, the re-ID performance will be higher if the encoded features are more representative.
Euclidean distance is used to calculate the extracted features after $l_2$-norm normalization \cite{PUL}.
As shown in the top rows of Table~\ref{tab:re-ID}, our BodyROI7 model achieves 0.338 and 0.355 (with PCA) rank-1 performance, higher than our WholeBody model, which is in accordance with the sampling results in Sec.~\ref{sec:sampling_comparison}. 
Besides, our method can achieve comparable performance with the unsupervised baseline methods, which indicates that our encoder can extract not only generative but also discriminative features.

\myparagraph{Using generated virtual image pairs.}
We use the generated image pairs to train the re-ID model and use the re-ID performance to evaluate our generation framework in an indirect manner. 
We first generate the VM re-ID dataset consisting of 500 identities with 24 images for each ID as illustrated in Fig.~\ref{fig:Paper_virtual_market}. 
For each identity, we randomly sample one foreground feature and 24 background features and randomly select 24 pose keypoint heatmaps from the Market-1501 training data.
Then, we use the same re-ID model and training procedure as in \cite{PUL}, but with different training data. 
As shown in the bottom rows of Table~\ref{tab:re-ID}, using our VM data the model can achieve the rank-1 performance 0.338 which is comparable to the model trained using another Duke re-ID dataset. 
When using the post-processing progressive unsupervised learning (PUL) proposed in \cite{PUL}, the rank-1 performance is improved to 0.369. 
Additionally, using our VM data, we can train a metric model, \eg KISSME~\cite{kissme}, and further improve the rank-1 performance to 0.375. 
Compared to the model trained using CUHK03 (rank-1 0.300) or Duke (rank-1 0.361) re-ID dataset with expensive human annotations, our method achieves better performance using only Market dataset without identity labels. 
These results show that our disentangled generated images are similar to the real data and can be further beneficial to re-ID tasks. 

\vspace{-0.5mm}
\section{Conclusion}
\vspace{-0.5mm}
\label{sec:conclusion}
We propose a novel two-stage pipeline for addressing the person image generation task. 
Stage-\RN{1} disentangles and encodes three modes of variation in the input image, namely foreground, background and pose, into embedding features then decodes them back to an image using a multi-branched reconstruction network. 
Stage-\RN{2} learns mapping functions in an adversarial manner for mapping noise distributions to feature embedding distributions guided by the decoders learned in stage-\RN{1}.
Experiments show that our method can manipulate the input foreground, background and pose, and sample new embedding features to generate intended manipulations of these factors, thus providing more control.
In the future, we plan to apply our method to faces and rigid object images with different types of structure.

\noindent {\bf Acknowledgments}
This research was supported in part by Toyota Motors Europe, German Research Foundation (DFG CRC 1223).
{\small
\bibliographystyle{ieee}
\bibliography{egbib}
}



\clearpage 
\noindent
{\Large {\textbf{Supplementary materials}}}
\\
\setcounter{section}{0}
\renewcommand\thesection{\Alph{section}}

This supplementary material includes additional details regarding the network architecture (\S\ref{sec:supp_arch}) and training (\S\ref{sec:supp_train}), as well as extended results for image manipulation (\S\ref{sec:supp_manipulation}), pose-guided person image generation (\S\ref{sec:supp_PG2}), inverse interpolation (\S\ref{sec:supp_interpolation}) and image sampling (\S\ref{sec:supp_sampling}), respectively.

\section{Network architecture}
\label{sec:supp_arch}

\begin{figure} [htp]
\scriptsize
\centering
  \includegraphics[width=0.7\linewidth]{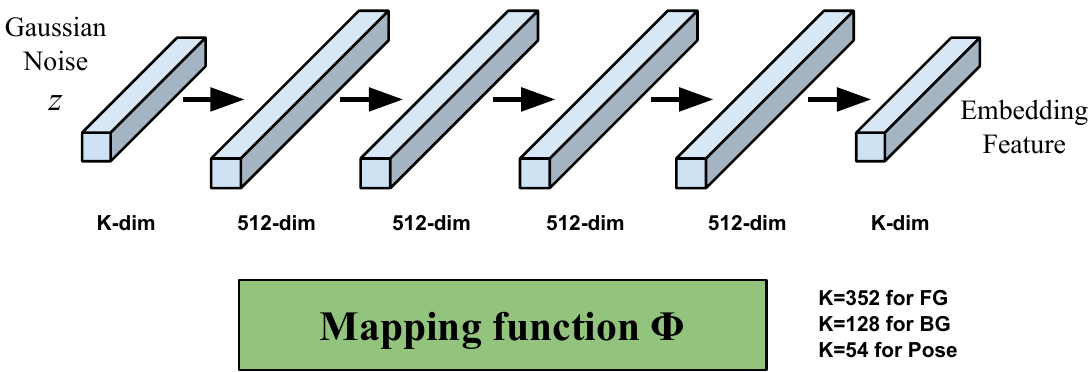}
\caption{Network architecture of the mapping functions for FG, BG and Pose in stage-\RN{2}.
}
\label{fig:supp_arch_stag2}
\end{figure}

\begin{figure*} [htp]
\scriptsize
\centering
\begin{minipage}{0.49\textwidth}
  \centering
  \includegraphics[width=1\linewidth]{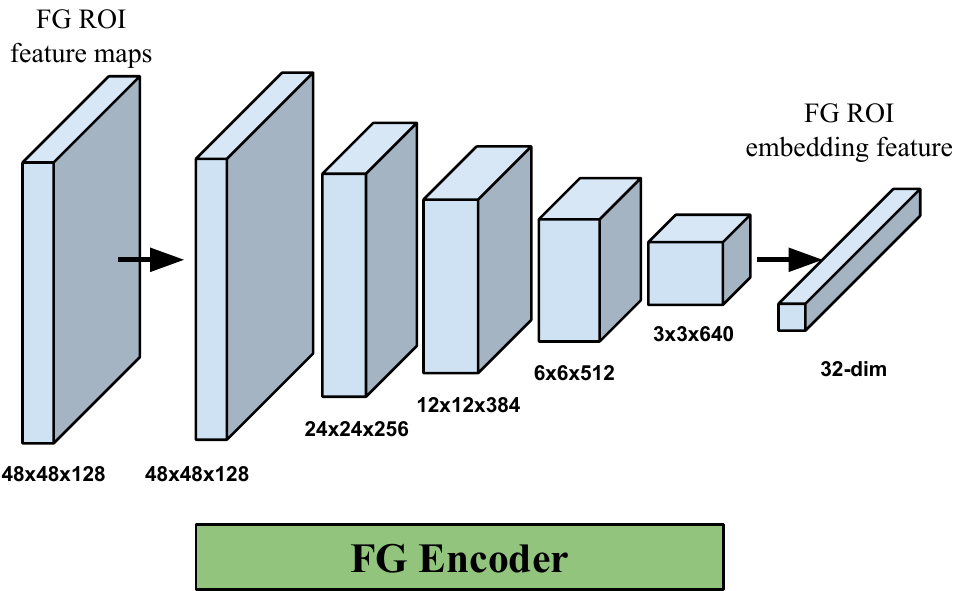}\\
(a)
\end{minipage}
\hfill
\begin{minipage}{0.49\textwidth}
  \centering
  \includegraphics[width=1\linewidth]{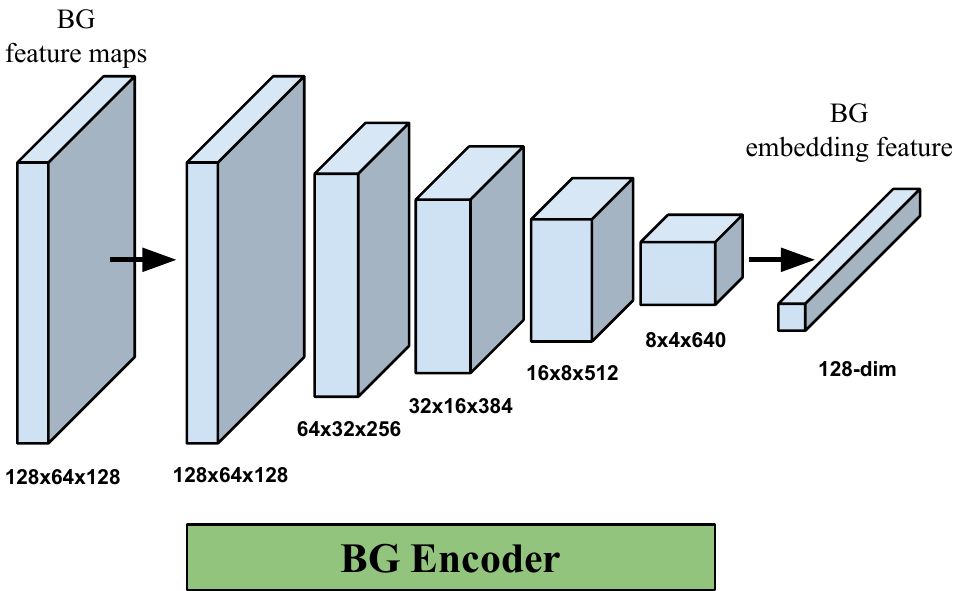}\\
(b)
\end{minipage}
\hfill
\begin{minipage}{0.99\textwidth}
  \centering
  \includegraphics[width=0.95\linewidth]{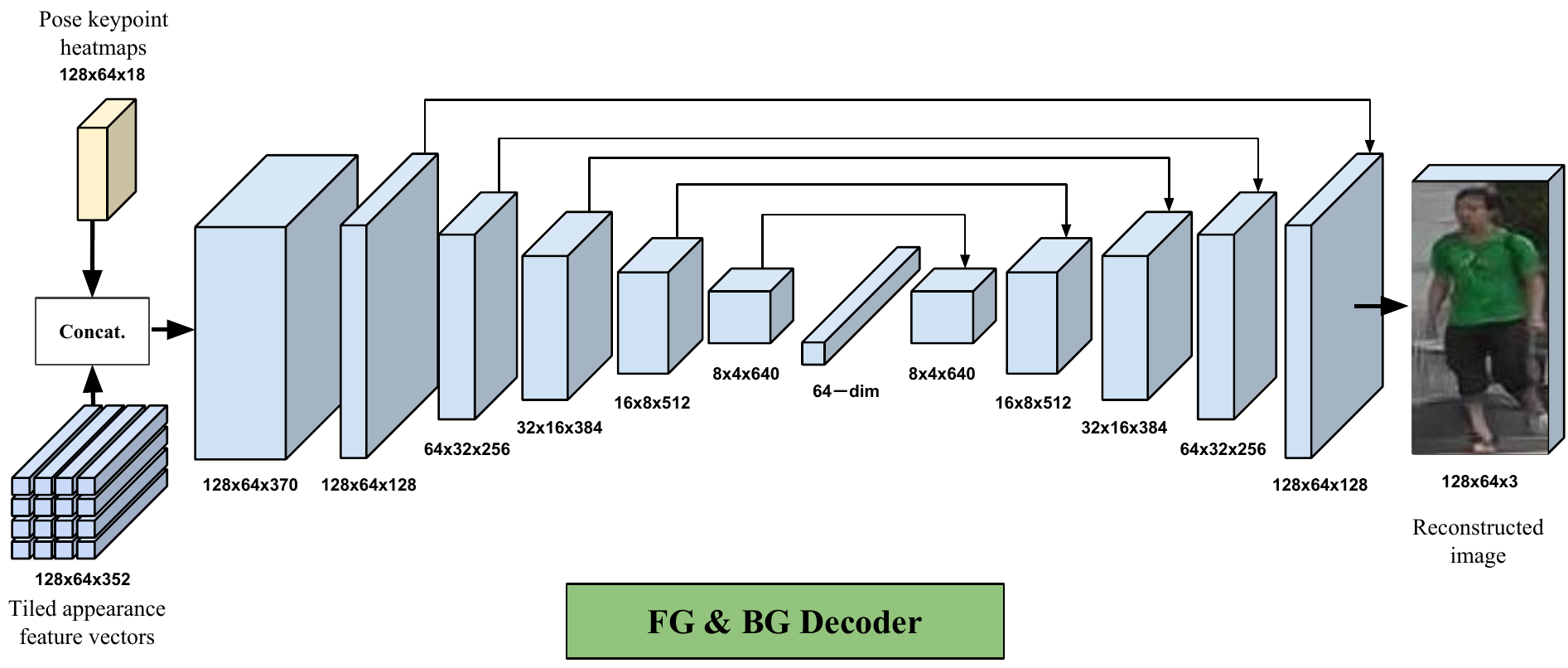}\\
(c)
\end{minipage}
\hfill
\begin{minipage}{0.99\textwidth}
  \centering
  \includegraphics[width=1\linewidth]{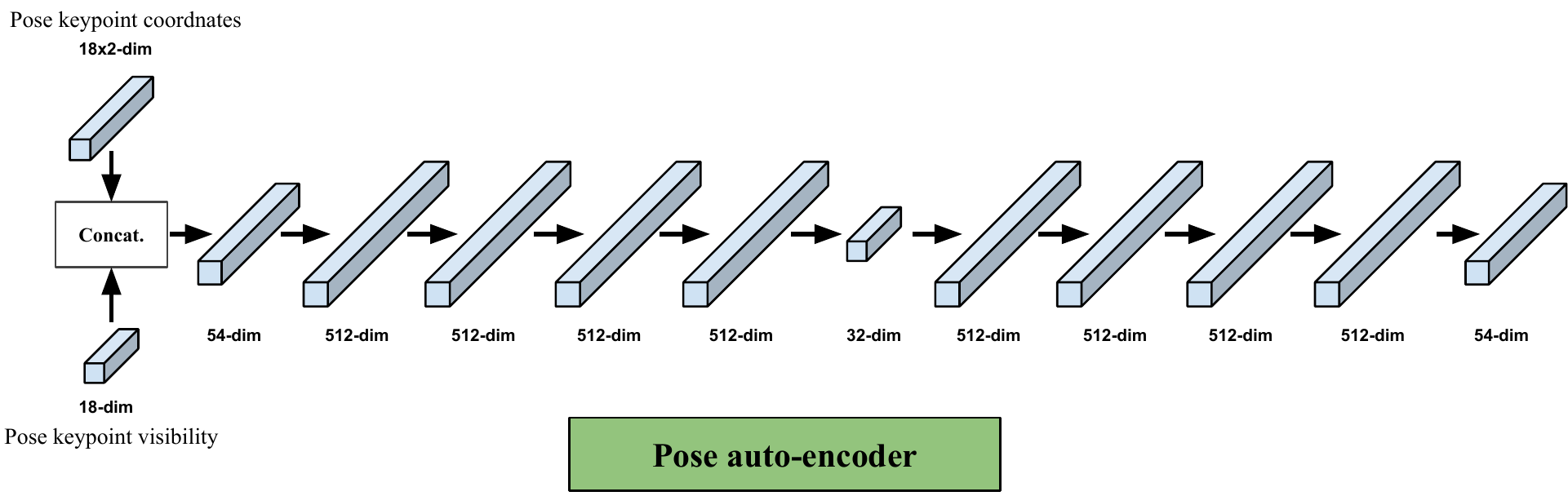}\\
(d)
\end{minipage}
\caption{Network architectures of stage-\RN{1}. (a) FG encoder, fed with the extracted 7 FG body ROI feature maps and outputting 7 FG embedding features of 32-dim after 5 convolutional residual blocks.
(b) BG encoder, fed with the BG feature maps and outputting a BG embedding feature of 128-dim after 5 convolutional residual blocks.
(c) FG and BG decoder, fed with the concatenated appearance and pose feature maps and outputting the generated image after the ``U-net''-based~\cite{U-net} architecture.
(d) Pose auto-encoder, fed with the concatenated keypoint coordinates and visibility vector and outputting the reconstructed vector after the auto-encoder.
}
\label{fig:supp_arch_stag1}
\end{figure*}

In this section, we provide details regarding the network architectures in our two-stage framework used on the Market-1501 dataset. Fig.~\ref{fig:supp_arch_stag1} shows 4 network architectures used at stage-\RN{1}: 1) FG encoder consists of 5 convolutional residual blocks; 2) BG encoder consists of 5 convolutional residual blocks; 3) FG \& BG decoder follows a ``U-net''-based architecture; 4) Pose auto-encoder follows a fully-connected auto-encoder architecture. 
Fig.~\ref{fig:supp_arch_stag2} shows the network architecture of the mapping functions $\Phi$ used at stage-\RN{2}. It contains 4 fully-connected residual modules.

\section{Training details}
\label{sec:supp_train}
On Market-1501, our method is applied to disentangle the image into three factors: foreground, background and pose.
We train the foreground and background models with a mini-batch of size $16$ for $\sim$70$k$ iterations at stage-\RN{1} and with a mini-batch of size $32$ for $\sim$30$k$ iterations at stage-\RN{2}. The pose models are trained with a mini-batch of size $64$ for $\sim$30$k$ iterations at stage-\RN{1} and with a mini-batch of size $32$ for $\sim$60$k$ iterations at stage-\RN{2}.

DeepFashion data contain clean background, therefore, our method is applied to disentangle the image into only two factors: appearance (\ie foreground) and pose.
We train the appearance model with a minibatch of size $6$ for $\sim$100$k$ iterations at stage-\RN{1} and with a minibatch of size $16$ for $\sim$60$k$  iterations at stage-\RN{2}. 
The pose models are trained with a minibatch of size $32$ for $\sim$30$k$  iterations at stage-\RN{1} and with a minibatch of size $32$ for $\sim$60$k$  iterations at stage-\RN{2}.

On both datasets, we use the Adam optimizer \cite{Adam} with weights $\beta_1=0.5$ and $\beta_2=0.999$. 
The initial learning rate is set to $2e$-$5$. For adversarial training, we optimize the discriminator and generator alternatively.

\section{Image manipulation results}
\label{sec:supp_manipulation}

\begin{figure*} [htp]
\scriptsize
  \centering
  \includegraphics[width=0.99\linewidth]{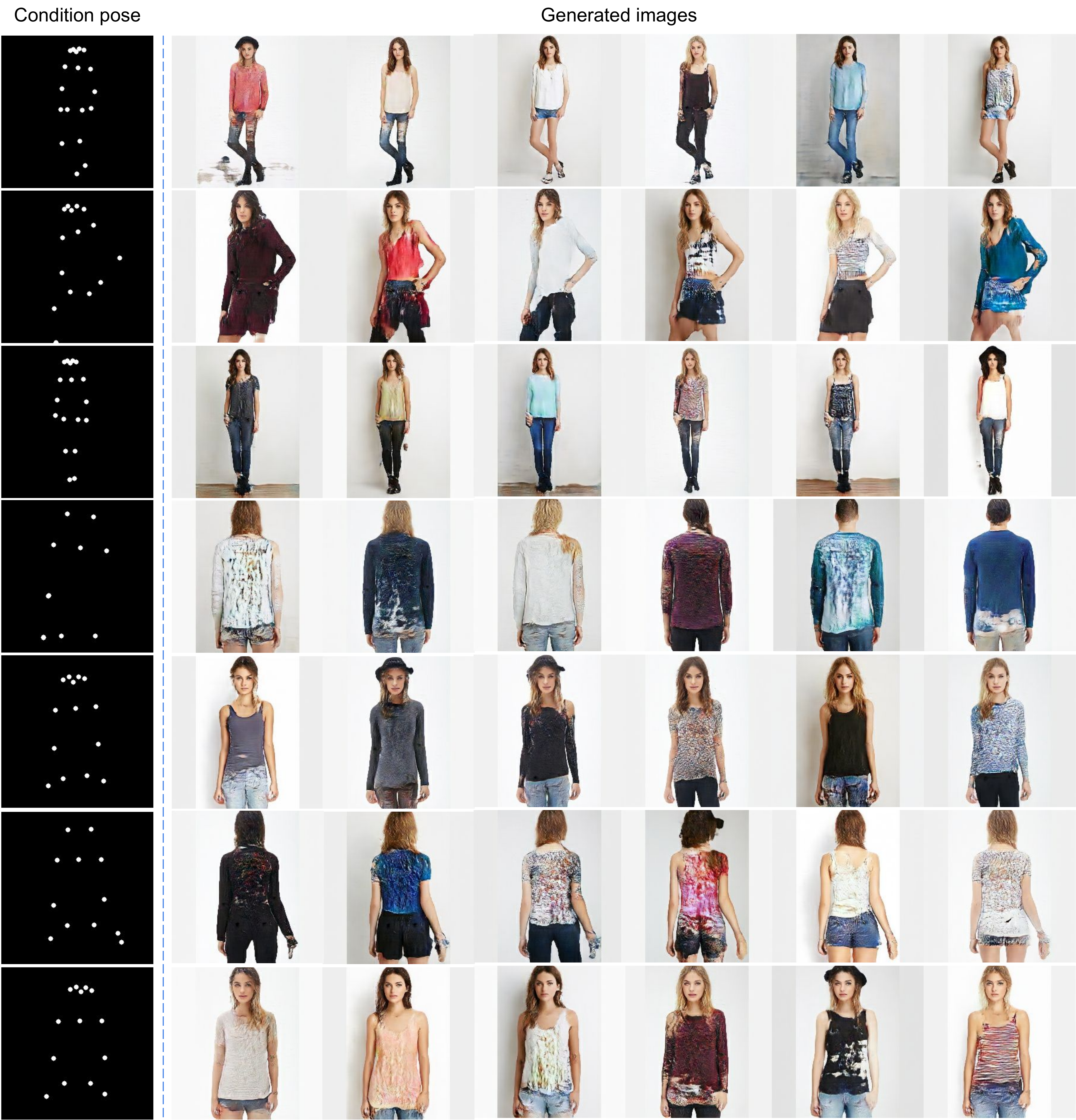}\\
  \caption{Appearance sampling (fixed Pose) results on the DeepFashion dataset. In each row, 6 different appearance factors are sampled from Gaussian noise and the pose factor is fixed to a real one.
  }
\label{fig:supp_DF_samplingApp}
\end{figure*}

\begin{figure*} [htp]
\scriptsize
  \centering
  \includegraphics[width=0.99\linewidth]{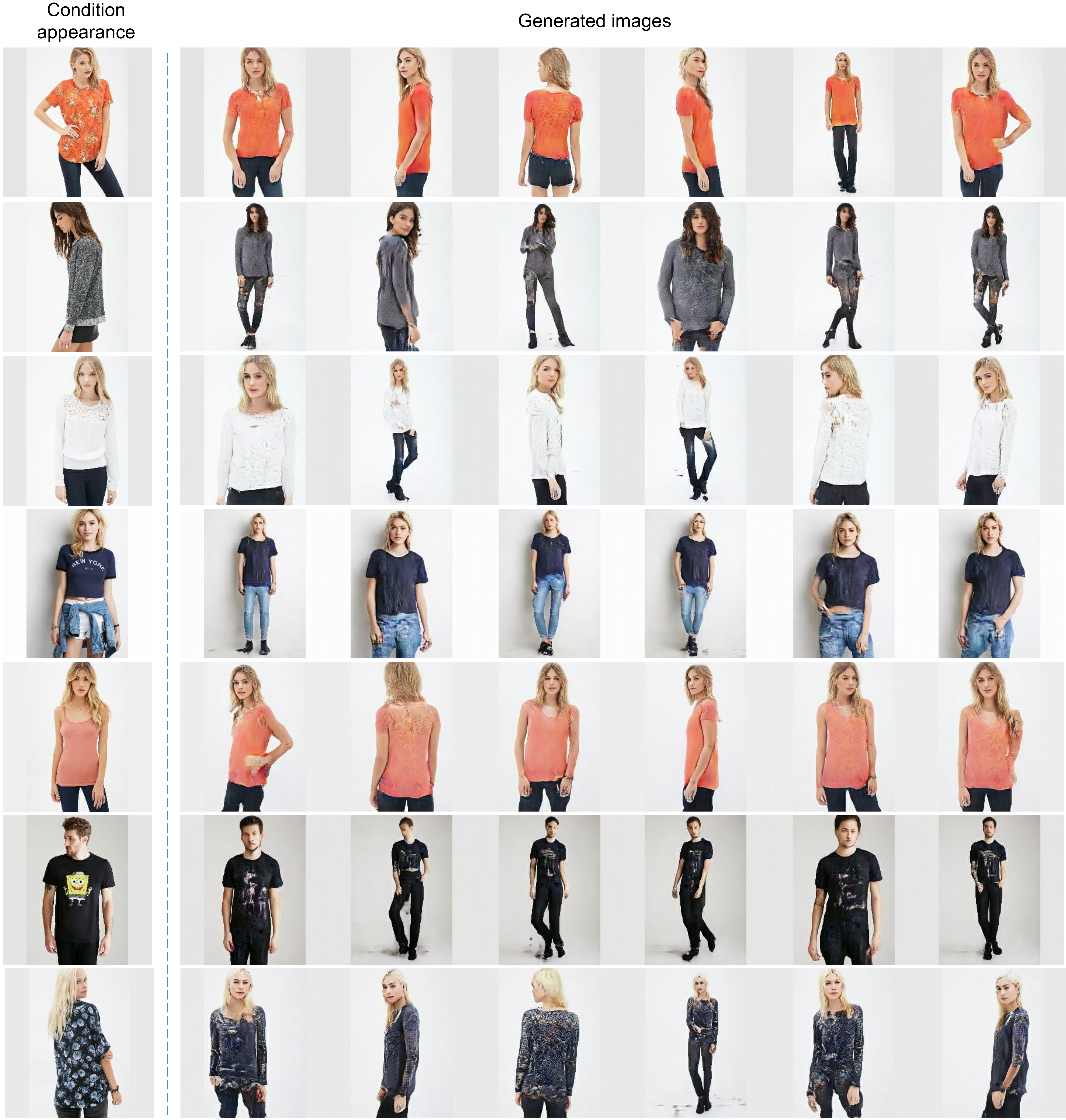}\\
  \caption{Pose sampling (fixed Appearance) results on the DeepFashion dataset. In each row, 6 different pose factors are sampled from Gaussian noises and the appearance factor is fixed to a real one.}
\label{fig:supp_DF_samplingPose}
\end{figure*}

In Fig.~\ref{fig:supp_DF_samplingApp} and Fig.~\ref{fig:supp_DF_samplingPose}, we provide results on appearance sampling and pose sampling for the DeepFashion dataset as an extension of Fig. 1 in the main paper. For each factor, we sample the embedding feature from Gaussian noise and fix the other factors by using the embedding feature extracted from the real data as explained in Sec. 4.2 in the main paper.

\section{Pose-guided person image generation results}
\label{sec:supp_PG2}
\begin{figure*} [htp]
\scriptsize
  \centering
  \includegraphics[width=0.75\linewidth]{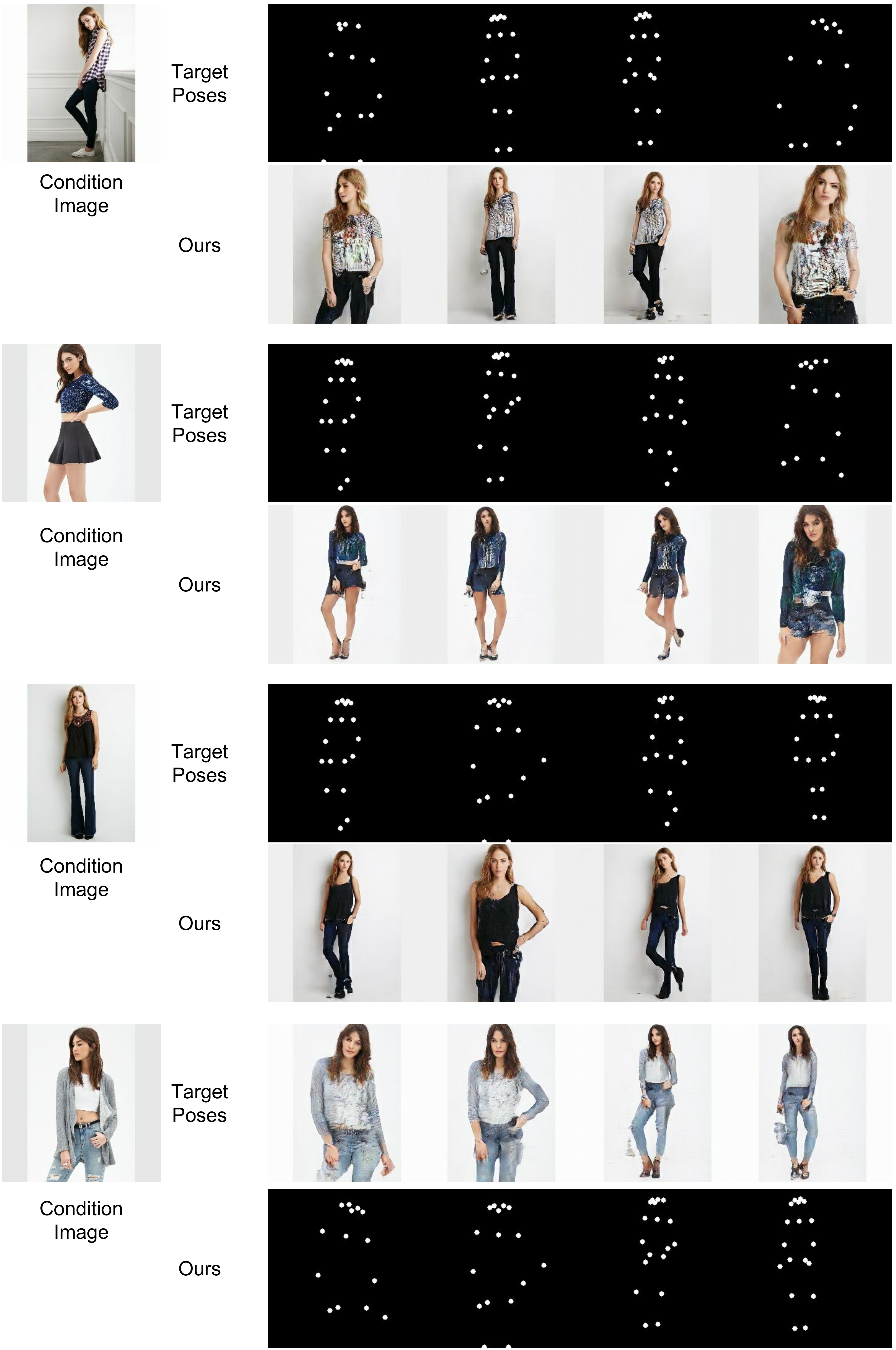}\\
  \caption{Generated results for one appearance with various poses on the DeepFashion dataset.}
\label{fig:supp_DF_seq}
\end{figure*}

For pose-guided person image generation, we provide more generated results. As an extension of Fig. 5 in the main paper, Fig.~\ref{fig:supp_DF_seq} shows the generated images of one appearance with various real poses selected randomly from DeepFashion. 

\section{Inverse interpolation results}
\label{sec:supp_interpolation}

\begin{figure*} [htp]
\scriptsize
\centering
\begin{minipage}{0.95\textwidth}
\hspace{-0.2cm}
  \centering
  \includegraphics[width=0.99\linewidth]{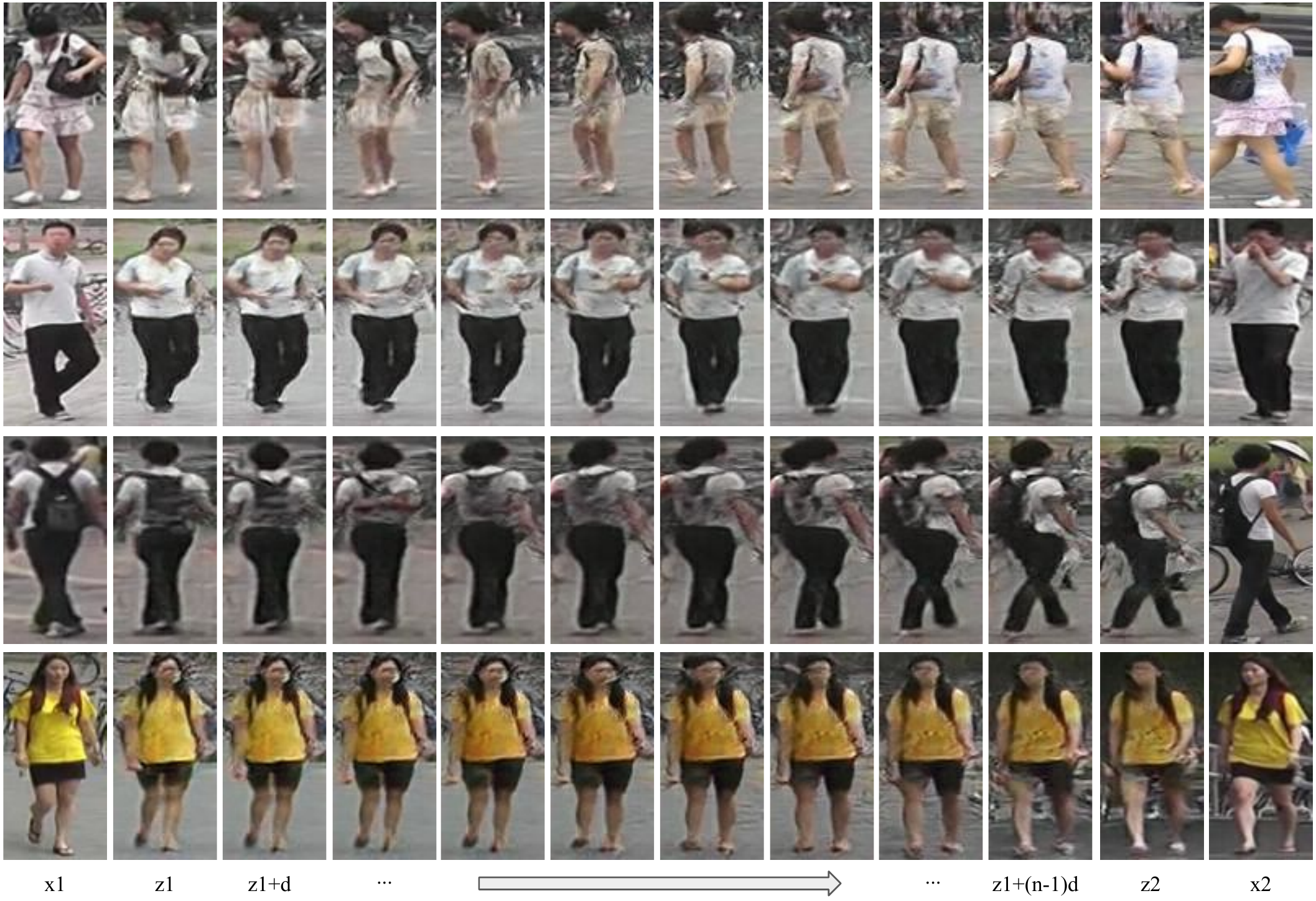}\\
(a)
\end{minipage}
\hfill
\begin{minipage}{0.95\textwidth}
  \centering
  \includegraphics[width=0.99\linewidth]{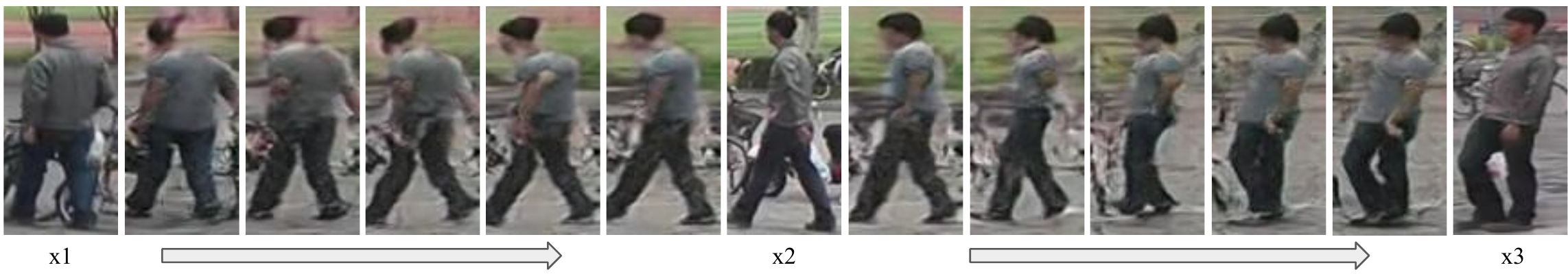}\\
(b)
\end{minipage}
\hfill
\begin{minipage}{0.95\textwidth}
  \centering
  \includegraphics[width=0.99\linewidth]{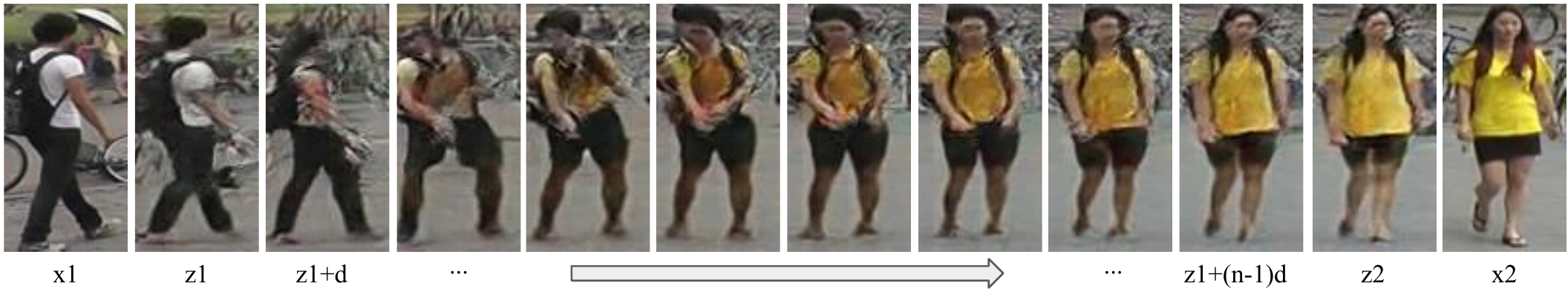}\\
(c)
\end{minipage}
\vspace{-0.2cm}
  \caption{Inverse interpolation results on Market-1501. (a) Interpolation between two images of the same person. (b) Interpolation between three images of the same person. (c) Interpolation between two images of different persons. }
\label{fig:Paper_interpolate_supp}
\end{figure*}

In this section, we provide more inverse interpolation results in Fig.\ref{fig:Paper_interpolate_supp} as an extension of Fig. 6 in the main paper. 
For two images $x_1$ and $x_2$, we find the corresponding Gaussian codes $z_1$ and $z_2$ as explained in the Sec. 4.4 of the main paper.
As shown in Fig.~\ref{fig:Paper_interpolate_supp}(a)(b), our method successfully generates the intermediate states between two images of the same person.
Note that, the inverse interpolation between two images of different persons is more challenging (see Fig.~\ref{fig:Paper_interpolate_supp}(c)) since we need to interpolate both the appearance and pose.

\section{Image sampling results}
\label{sec:supp_sampling}

\begin{figure*} [htp]
\scriptsize
\begin{minipage}{0.33\textwidth}
\centering
  \centering
  \includegraphics[width=0.99\linewidth]{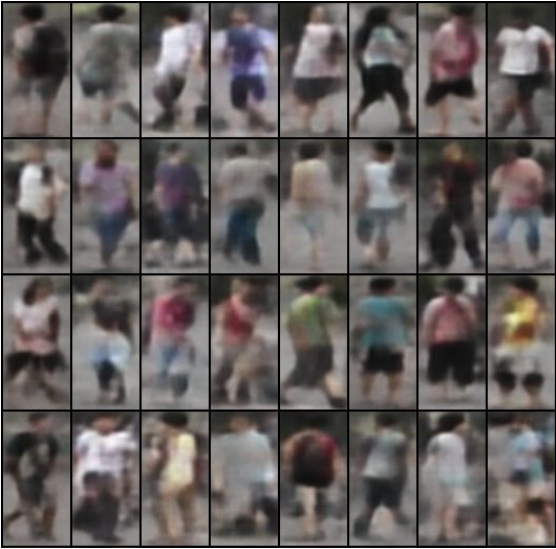}\\
(a) Vanilla VAE
\vspace{1mm}
\end{minipage}
\hfill
\begin{minipage}{0.33\textwidth}
\centering
  \centering
  \includegraphics[width=0.99\linewidth]{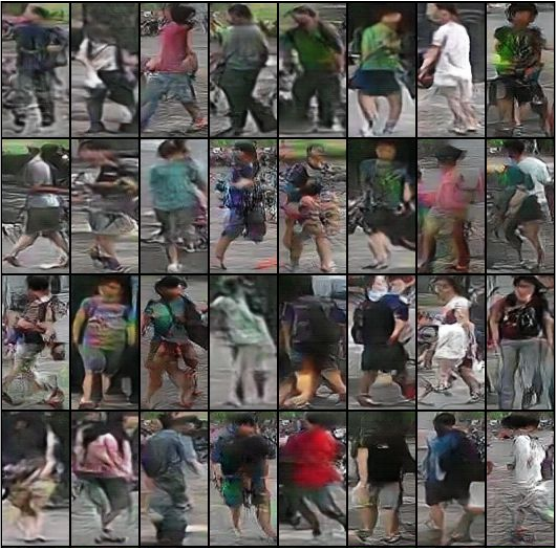}\\
(b) Vanilla DCGAN
\vspace{1mm}
\end{minipage}
\hfill
\begin{minipage}{0.33\textwidth}
\centering
  \centering
  \includegraphics[width=0.99\linewidth]{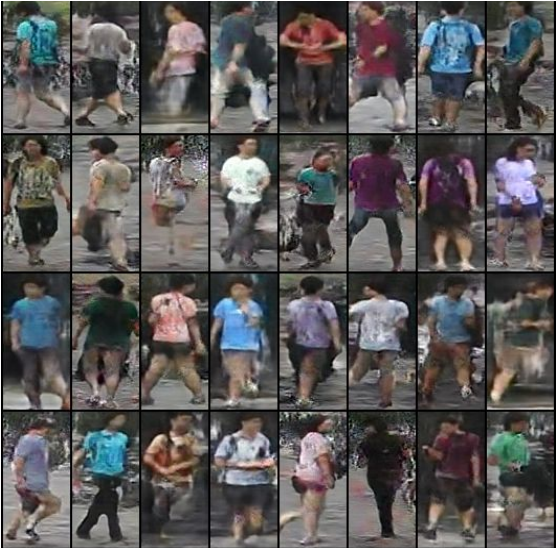}\\
(c) Ours - Whole Body
\vspace{1mm}
\end{minipage}
\hfill
\begin{minipage}{0.33\textwidth}
\centering
  \centering
  \includegraphics[width=0.99\linewidth]{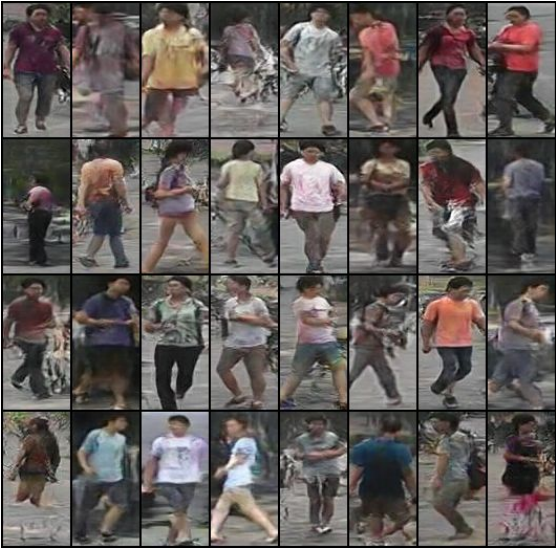}\\
(d) Ours - BodyROI7
\end{minipage}
\hfill
\begin{minipage}{0.33\textwidth}
\centering
  \centering
  \includegraphics[width=0.99\linewidth]{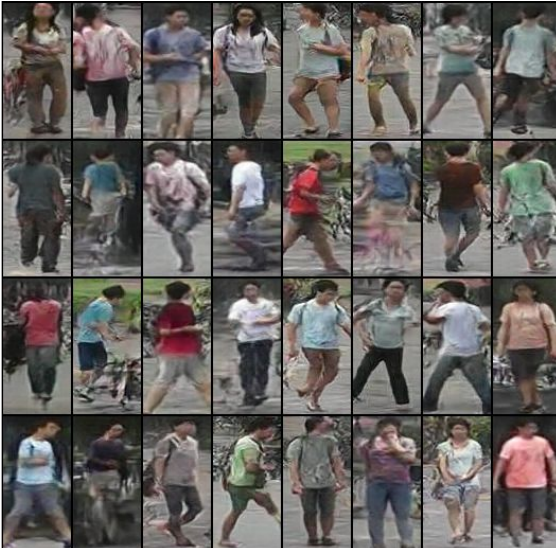}\\
(e) Ours - BodyROI7 with real pose from training set
\end{minipage}
\hfill
\begin{minipage}{0.33\textwidth}
\centering
  \centering
  \includegraphics[width=0.99\linewidth]{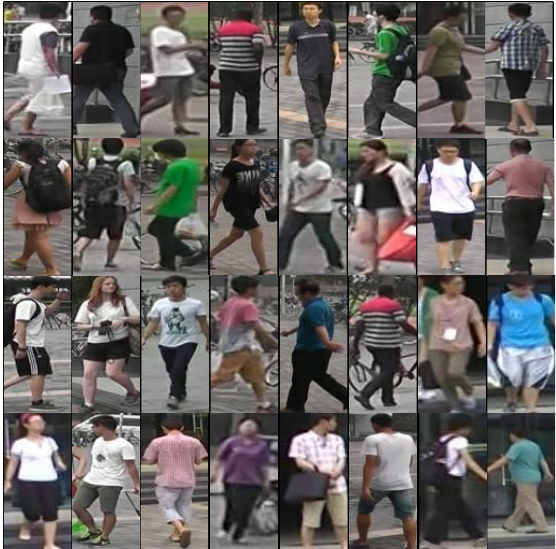}\\
(f) Real data
\end{minipage}
\caption{Sampling results. (a) Vanilla VAE; (b) Vanilla DCGAN; (c) Ours - Whole Body; (d) Ours - BodyROI7; (e) Ours - BodyROI7 pose with real pose from training set; (f) Real data. }
\label{fig:supp_Market_sampling}
\end{figure*}

We also give more sampling results as extensions of Fig.7 in the main paper. Fig.~\ref{fig:supp_Market_sampling} shows the sampling results (a-e) and real images (f) on Market-1501 dataset.
VAE generates blurry images and DCGAN sharp but unrealistic person images. In contrast, our model generates more realistic images (c)(d)(e). By comparing (d) and (c), we observe that our model using body ROI generates more sharp and realistic images whose colors on each body part are more natural. By comparing (e) and (d), we see that when sampling foreground and background but using the real pose keypoints randomly selected from the training data, we generate better results.

\end{document}